%% file: HigherOrderMPNN.tex
\documentclass[twoside]{article}
\usepackage[accepted]{aistats2024}

\input{math_commands.tex}

\usepackage{hyperref}
\usepackage{url}
\usepackage[utf8]{inputenc} 
\usepackage[T1]{fontenc}    
\usepackage{hyperref}       
\usepackage{booktabs}       
\usepackage{nicefrac}       
\usepackage[protrusion=true,expansion=false]{microtype}      
\usepackage{xcolor}         
\usepackage{subfig}

\usepackage{url,amsthm,amsmath,amssymb,amsfonts}
\usepackage{tikz}
\usepackage{bbm}
\usepackage{rbase} 
\usepackage{paralist}
\usepackage{enumitem}
\definecolor{hcitecolor}{RGB}{40,40,160}
\usepackage{mdframed}
\usepackage{pbox}
\usepackage{ifthen}
\newcommand{\ifdraft}[1]{}

\usetikzlibrary{shadows}
\usetikzlibrary{matrix, calc, positioning, arrows,shapes,backgrounds, arrows.meta, quotes, patterns}
\usetikzlibrary{shapes.geometric,fit,matrix,positioning,shapes.multipart}
\usetikzlibrary{decorations.markings}
\usetikzlibrary{decorations.pathreplacing,fadings}
\pgfdeclarelayer{background}
\pgfsetlayers{background,main}
\usetikzlibrary{topaths, calc,3d}
\usetikzlibrary{fit}

\newtheorem{theorem}{Theorem}

\newtheorem{prop}[theorem]{Proposition}

\newtheorem{defn}{Definition}

\newenvironment{pfof}[1]{\textbf{Proof of #1.}}{\mbox{}\hfill\m{\blacksquare}\\ \mbox{}\noindent}

\newcommand{\Fin}{T^{\textrm{in}}}
\newcommand{\Fout}{T^{\textrm{out}}}
\newcommand{\Ttin}{T^{\textrm{in}}}
\newcommand{\Tout}{T^{\textrm{out}}}

\newcommand{\dcap}{d^{\cap}}

\ignore{
\makeatletter
\def\thm@space@setup{%
  \thm@preskip=6pt plus 0pt minus 0pt
  \thm@postskip=0pt plus 0pt minus 0pt 
}
\makeatother
}

\newcommand{\includepicc}[2]{\centerline{\includegraphics[width=#1\textwidth]{#2}}}

\newcommand{\neuron}{\mathfrak{n}}

\newcommand{\sm}[1]{\m{\smash{#1}}}

\newcommand{\dn}{{\downarrow}}

\renewcommand{\v}{\ensuremath{\textbf{\textrm{v}}}}

\newcommand{\seteq}{\mathop{\stackrel{\textrm{set}}{=}}}
\newcommand{\Ptensor}{\m{P}-tensor}
\newcommand{\Ptensors}{\m{P}-tensors}
\newcommand{\Ttens}{T}
\newcommand{\Ts}{T^{\textrm{src}}}
\newcommand{\Td}{T^{\textrm{dest}}}
\newcommand{\tsum}{\textstyle \sum}

\makeatletter
\providecommand*{\cupdotB}{\mathbin{\mathpalette\@cupdotB{}}}
\newcommand*{\@cupdotB}[2]{\ooalign{$\m@th#1\cup$\cr\hidewidth$\m@th#1\cdot$\hidewidth}}
\makeatother

\usepackage{fancyhdr}

\usepackage[round]{natbib}
\renewcommand{\bibname}{References}
\renewcommand{\bibsection}{\subsubsection*{\bibname}}
\begin{document}

\twocolumn[

\aistatstitle{P-tensors: a General Framework for Higher Order Message Passing in Subgraph Neural Networks}

\aistatsauthor{Andrew Hands \And Tianyi Sun \And Risi Kondor }

\aistatsaddress{Department of Computer Science\\ University of Chicago\\ \texttt{hands@uchicago.edu}\And 
Computational and Applied Math\\Department of Statistics\\University of Chicago\\ \texttt{tianyisun@uchicago.edu} \And 
Department of Computer Science\\Department of Statistics\\University of Chicago\\ \texttt{risi@uchicago.edu} }
]

\input{abstract}
\input{intro}
\input{background2}

\input{equivariance}
\input{ptensors}
\input{gnns}

\input{results}
\input{conclusions}
\bibliography{bibliography,gnn1}
\vspace{-10pt}
\input{checklist}
\clearpage 
\onecolumn
\aistatstitle{ Appendix }
\input{proofs2}
\input{appendix-experimental-details}
\end{document}

%% file: math_commands.tex
\usepackage{amsmath,amsfonts,bm}

\def\eqref#1{equation~\ref{#1}}

\def\1{\bm{1}}

\def\rf{{\textnormal{f}}}

\DeclareMathAlphabet{\mathsfit}{\encodingdefault}{\sfdefault}{m}{sl}
\SetMathAlphabet{\mathsfit}{bold}{\encodingdefault}{\sfdefault}{bx}{n}



%% file: abstract.tex
\begin{abstract}
Several recent papers have proposed increasing the expressive power of graph neural networks 
by exploiting subgraphs or other topological structures. In parallel, researchers 
have investigated higher order permutation equivariant networks.  
In this paper we tie these two threads together by providing a general framework for higher order 
permutation equivariant message passing in subgraph neural networks. 
In this paper we introduce a new type of mathematical object called 
$P$-tensors, which provide a simple way to define the most general form of 
permutation equivariant message passing in both the above two categories of networks.  
We show that the \m{P}-tensors paradigm can achieve state-of-the-art performance on benchmark molecular datasets.
\end{abstract}

%% file: intro.tex
\section{INTRODUCTION} 

Graph Neural Networks (GNNs) have proved to be remarkably successful in a wide range of domains 
from filling in edges in social networks to predicting the chemical properties of molecules. 
Amongst graph neural networks, so-called Message Passing Neural Nets (MPNNs), which loosely imitate 
convolution on the graph, are particularly popular and have found 
use in almost every branch of science \citep{gilmer2017neural}. 
However, it has also been shown that there are severe theoretical limitations 
on the expressive power of MPNNs,  
and that in some cases they fail to capture even relatively simple topological features 
such as cycles \citep{morris2019weisfeiler,xu2018powerful}.  

To develop more expressive GNNs, researchers have developed models that consider higher order representations 
of graphs, such as Invariant Graph Neural Networks (IGNs) \citep{maron2018,maron2019provably,pmlr-v97-maron19a}, 
and models that explicitly account for specific types of subgraphs  
\citep{frasca2022understanding,pmlr-v139-bodnar21a}.  
In the first category of models, at the cost of increased time and space complexity, 
in principle, one can use arbitrarily high order permutation equivariant representations 
\citep{maron2019provably,pmlr-v97-maron19a}.  
In the second category one can encode a rich set of substructures 
allowing the network to learn different sets of weights for each type of structure 
\citep{bodnar2021weisfeiler}.  
However, the question of how to bring these two lines of research together and design a general 
architecture where neurons corresponding to varied substructures can pass messages 
to each other in the most general possible way has 
so far been elusive.  


\textbf{Main contributions.} 
In this paper we present a general framework for extending the message passing paradigm to a 
settting where (a) the titular ``neurons'' can correspond to not just individual vertices, 
but specific types of subgraphs or other topological structures related to the underlying graph, as 
long as these structures are selected by a permutation invariant selection policy; (b) the neurons 
communicate with each other via \emph{higher order} message passing  
in the sense that with respect to local permutations 
the messages behave as permutation covariant vectors, matrices or tensors. 

Our first technical contribution is to give a precise definition of equivariance in this setting,  
which, to the best of our knowledge has not been previously done 
(Definitions \ref{def: covariant HOSNN}--\ref{def: equivariant HOSNN}).
Next, we introduce \m{P}-tensors, a new mathematical device that makes it easy to decribe 
and implement higher order message passing between subgraphs. 
Here we have two technical results. 
First we derive the form of all possible equivariant linear maps between \m{P}-tensors, which turns 
out to be a generalization of the results of \citep{maron2018}, but affording a larger set of possible  
interactions (Theorem \ref{thm: main theorem}). Second, we show that any subgraph neural network 
utilizing these maps is equivariant in the sense of our earlier definition (Theorem \ref{thm: equivariance}).
In our experiments we find that the flexibility and generality of the \Ptensors{} framework 
does indeed pay off, in particular, on the  ZINC 12K benchmark we reach 
state-of-the-art results. 




%% file: background2.tex
\section{BACKGROUND: MESSAGE PASSING NEURAL NETWORKS} 

Let \m{\Gcal} be an undirected graph with \m{n} vertices and \m{A\tin\RR^{n\times n}} be its adjacency matrix. 
Graph neural networks (GNNs) learn a function \m{\Phi\colon A_\Gcal\mapsto \Phi(A_\Gcal)} 
embedding such graphs in some Euclidean space \m{\RR^D}.  
The fundamental constraint on \m{\Phi} is that it must be invariant to relabeling the vertices.  
If we change the order in which the vertices are numbered by a permutation 
\m{\sigma}, the adjacency matrix transforms as 
\begin{equation}\label{eq: Aact}
A\mapsto A',\qquad \textrm{with}\qquad
[A']_{i,j}=A_{\sigma^{-1}(i), \sigma^{-1}(j)}. 
\end{equation}
However, \m{A'} still represents the same graph \m{\Gcal}, so overall the network must satisfy \m{\Phi(A)\<=\Phi(A')}.

Requiring permutation invariance at the level of individual neurons would be too restrictive. 
Instead, modern GNNs are designed in such a way that their internal layers are \emph{equivariant} 
(rather than invariant) to permutations,   
meaning that under \rf{eq: Aact} the outputs of the layers do change, 
but do so in a specific, controlled way. 
The last layer of the network is designed to cancel out these transformations, typically by pooling over the vertices, 
and thus guarantees that the final output is invariant. 

Presently the most popular approach to building equivariant GNNs 
is the \emph{Message Passing Neural Network} (MPNN) paradigm 
\citep{gilmer2017neural}, where 
the titular ``neurons'' are 
attached to the vertices of the graph and 
communicate with each other by sending messages to their neighbors.  
In the simplest case, the output of the neuron at vertex \m{i} in layer \m{\ell} is a vector 
\m{f_i^\ell\tin \RR^{D_\ell}}, and the update rule is 
\begin{equation}\label{eq: MPNN-update}
f^\ell_i=\eta\brBig{W_\ell\! \sum_{j\in\Ncal(i) } f^{\ell-1}_j+b^\ell_i}. 
\end{equation}
Here \m{\Ncal(i)} denotes the neighbors of node \m{i}, \m{W_\ell} is a learnable weight matrix, 
\m{b^\ell_i} is a learnable bias term and \m{\eta} is a pointwise non-linearity. 

It is easy to see that under the transformation \rf{eq: Aact}, the output of each neuron 
in such an MPNN becomes \sm{{f'}^\ell_i=f^\ell_{\sigma^{-1}(i)}}, and 
in this sense the network as a whole is permutation equivariant. 
The message passing process also bears some similiarity to classical convolution in e.g., image 
processing, and captures the intuitive idea that information in complex networks should propagate 
via local connections. 

Despite these attractive properties, classical MPNN do have significant limitations, 
the most obvious of which is that 
each vertex \m{i} just \emph{sums} all the incoming messages from its neighbors. 
The associative and commutative nature of summation is  critical for ensuring equivariance, but it also makes 
MPNNs myopic in the sense that once the activations have been summed, 
downstream neurons have no way to 
distinguish between which part of the incoming message came from which neighbor  
\citep{CCN-JCP,morris2019weisfeiler,xu2018powerful,chen2020can,you2021identity}. 
From a theoretical point of view, the consequence is that classical MPNNs are only 
as powerful as the first order 
Weisfeiler-Leman test \citep{xu2018powerful,weisfeiler1968reduction}. 
In response to these criticisms, the community has been exploring various ways to 
generalize the message passing idiom. 

\subsection{Higher order equivariance}

One way to make MPNNs more expressive is to design architectures that 
are equivariant to higher order actions of the group of permutations, technically called the 
symmetric group, \m{\Sn}.  
Studying this problem at the most general level involves considering equivariance to each irreducible 
representation of \m{\Sn}, a rich but mathematically involved subject 
\citep{Sagan,Sannai2019universal,KerivenPeyre2019,thiede2020general}. 
For practical GNNs however it is usually sufficient to consider \m{k}'th order equivariance 
in the sense of how \m{\Sn} acts on \m{k}'th order tensors:
\begin{equation}
T\stackrel{\sigma}{\longmapsto}T'\qquad
[T']_{\sseq{i}{k}}=[T]_{\sigma^{-1}(i_1),\ldots,\sigma^{-1}(i_k)}.
\end{equation}
\citet{maron2018} derived the general form of linear neural network layers that are invariant to this 
action and showed that for higher values of \m{k} it captures 
much richer interactions than simple first-order message passing, which is essentially what 
classical MPNNs do. 
The fundamental limitation of this approach however is that the size of the tensor \m{T} grows exponentially with \m{k}. 
Therefore, even for moderate sized graphs, considering more than second or third order permutation 
equivariance becomes infeasible. 

\subsection{Subgraph neural networks}

The other natural way to increase the expressiveness of graph neural networks is to extend the 
MPNN paradigm to passing messages to/from edges and other subgraphs 
\citep{alsentzer2020subgraph,thiede2021,bevilacqua2022equivariant,frasca2022understanding}.  
This class of approaches is also attractive because in many applications 
subgraphs have explicit semantic meaning. In organic chemistry, for example, they 
can correspond to functional groups. 

At the extreme, subgraph neural networks allow combining two (or more) separate GNN algorithms 
in a recursive fashion, with one GNN running at the level of the subgraphs, and the other one combining their 
results at the global level. 
Subgraph neural networks are also related to hypergraph neural nets and simplicial 
complex networks \citep{Feng2019,dong2020hnhn,Ebli2020,pmlr-v139-bodnar21a,zhao2021stars,frasca2022understanding}. 

There is a natural connection between subgraph networks and higher order permutation equivariance, 
which is that to capture the underlying combinatorial structures, subgraph neurons cannot just communicate via  
scalar (invariant) messages. Rather, the messages must be indexed by the vertices of the sending 
and receiving subgraph, i.e., they must be permutation covariant objects. 
This  complicates deriving the rules of equivariant message passing, and 
most existing architectures employ somewhat ad hoc solutions.
The goal of the present paper is to combine the higher order message passing and subgraph neural network 
frameworks and derive the general laws of permutation equivariant message passing between subgraphs 
in higher order subgraph networks (HOSNNs). 

\ignore{

However, the more complicated the subgraph structure becomes, the more nontrivial it is to design 
an appropriate message passing scheme between the various constituents of the network, 
that can still guarantee global permutation equivariance. 
Treating the output of each subgraph neuron as a scalar and using a simple generalization of 
\rf{eq: MPNN-update} is  option, but more powerful networks that can truly exploit the 

but such a simplistic approach to some extent defeats the 
purpose of subgraph neurons. 

Instead, to fully harness the the power of subggraph neurons, 
the message passing process should reflect the precise way that the the sending and receiving neuron's 
subgraphs are related to each other. 

In this paper we study this problem 
}

\ignore{
several authors have proposed variations on the message passing 
paradigm where individual neurons correspond 
to \emph{sets} of vertices, typically, \emph{subgraphs}. 
Such \emph{subgraph-neurons} have the ability to represent information relating to the subgraph as a whole, 
almost as a ``GNN within a GNN''. 
While 
including relations between the vertices of the subgraph, 
and the output of the subgraph-neuron is a vector, matrix or tensor that is covariant 
with respect to \emph{internal} permutations of the vertices of just the subgraph.
Architectures of this type we call \emph{higher order message passing networks} (HO-MPNNs). 

Discussing the equivariance properties of HO-MPNNs is more challenging, because it 
involves the action of permutations at two distinct levels: 
on the level of the so-called \emph{subgraph selection policy} used to select the subgraphs, and 
on the level of the individual subgraph-neurons. 
In particular, it is the interaction between these two levels that is critical for global equivariance. 
In the following section we lay down a general framework for equivariant message passing in 
HO-MPNNs, and in Section \ref{sec: ptensors} we introduce P-tensors, a 
specific mathematical object used to construct such architectures. 
}

\ignore{
Correspondingly, the output of such subgraph-neurons is a vector, matrix or tensor that is covariant 
with respect to \emph{internal} permutations of the vertices of the subgraph. 
For example, if the subgraph in question 
is a triangle consisting of vertices \m{(i_1,i_2,i_3)}, the output might be a 3-index quantity 
\m{T\in \RR^{3\times 3\times C}}. 
Architectures of this type we call \emph{higher order message passing networks}. 

To describe the equivariance properties of higher order MPNNs, we need to examine not just the internal 
workings of the subgraph-neurons, but also the so-called \emph{selection policy} \m{\psi} used to select 
which subgraphs we apply the subgraph-neurons to. Typically, \m{\psi} is a function 
\m{A\mapsto \mathbb{D}} or \m{(A,L)\mapsto \mathbb{D}}, where \m{A} is the adjacency matrix, 
\m{L} is a matrix of input features (such as atom types) and \m{\mathbb{D}} is a set of subsets 
of vertices. In particular, in this paper we consider \emph{invariant} selection policies. 

\begin{defn}
\end{defn}

In such networks 
equivariance comes into play at two separate levels: globally, 
at the level of the subgraphs corresponding to different neurons being permuted with each other, 
and locally, at the level of the vertices belonging to an individual subgraph being permuted 
amongst themselves. In this paper we develop a general mathematical formalism for describing 
this situation an derive the most general form of message passing operations that 
ensures overall equivariance (Figure \ref{fig: 2-level}). 

\input{fig-2level}
}

%% file: fig-2level.tex
\begin{figure}
\centering
\includegraphics{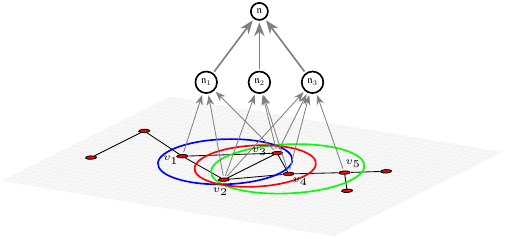}
\caption{
\label{fig: 2-level} 
In higher order message passing networks, permutation equivariance acts on two levels: 
(a) it affects which vertices a given neuron \m{\neuron} is responsible for (the referenced domain \m{\Dcal_\neuron}), and 
(b) it changes the ordering of the vertices in \m{\Dcal_\neuron} itself. 
Equivariance with respect to the first transformation is relatively easy to achieve, simply by ensuring that 
the policy used to select the reference domain is itself permutation covariant \cite{frasca2022understanding}. 
The second type of transformation is more subtle and requires the P-tensor formalism to account for. 
}
\end{figure}

%% file: equivariance.tex
\section{EQUIVARIANCE IN HIGHER ORDER SUBGRAPH NEURAL NETWORKS}\label{sec: equivariance}

The symmetric group acts on subgraph neural networks on two distinct levels: 
the vertices inside individual subgraphs get permuted, and the subgraphs themselves 
are permuted with each other. 
Our first challenge is to define what equivariance even means in this setting. 
We begin by formalizing how the subgraphs are selected. 


\begin{defn}\textbf{(Selection policy)}
Let \m{\Gcal} be a graph with vertex set \m{V=\cbrN{1,\ldots,n}}, adjacency matrix \m{A\tin\RR^{n\times n}} 
and optionally an input feature matrix \m{L\tin\RR^{n\times d}}.  
A subgraph selection policy is a function \m{\psi\colon A\mapsto \mathbb{D}} or 
\m{\psi\colon (A,L)\mapsto \mathbb{D}} where \m{\mathbb{D}} is a set of subsets of \m{V}. 
\end{defn}

The purpose of \m{\mathbb{D}} is to demark the subgraphs to which we assign higher order neurons. 
For example, we might define \m{\psi} to return the set of all edges in \m{\Gcal}, 
all paths of a given length, or all cycles. 
Each of these selection policies is \emph{invariant} in the following sense. 
\input{fig-subgraphmp}

\begin{defn}\textbf{(Invariant selection policy)} 
Let \m{\Gcal}, \m{A} and \m{L} be as above. Assume that under permuting the vertices 
\sm{A\mapsto A^{(\sigma)}} and \sm{L\mapsto L^{(\sigma)}} with 
\[A^{(\sigma)}_{i,j}=A_{\sigma^{-1}(i),\sigma^{-1}(j)}  \qquad\text{and}\qquad  
L^{(\sigma)}_{i,c}=L_{\sigma^{-1}(i),c}.\] 
Then a selection policy \sm{\psi\colon (A,L)\mapsto\mathbb{D}} is said to be invariant if 
for any permutation \m{\sigma\tin\Sn} and any selected set \sm{\cbrN{\sseq{i}{p}}\tin \psi(A,L)}, 
we have \sm{\cbrN{\sigma(i_1),\ldots,\sigma(i_p)}\tin \psi(A^{(\sigma)},L^{(\sigma)})}. 
\end{defn}

An invariant selection policy is a policy that depends only on the graph topology and the 
input vertex features, but not the (arbitrary) ordering of the vertices.  
All the subgraph selection policies considered in this paper are invariant policies.

\subsection{Two-level equivariance}
 
The output of each layer of a higher order subgraph neural network 
is a collection of vectors, matrices or tensors \sm{\Fcal=\cbrN{\Ttens_1,\ldots,\Ttens_m}} 
corresponding to the subgraphs 
\sm{\Scal_{\cbrN{i^1_{1}\,\ldots,i^1_{p_1}}},\ldots, \Scal_{\cbrN{i^m_{1}\,\ldots,i^m_{p_m}}}}
induced by the vertex sets picked out by the selection policy \m{\psi(A,L)}. 
The difficulty with this setup is two-fold:
\begin{compactenum}[1.]
\item Since \m{\psi} picks out subgraphs in an arbitrary order, the order in which the 
\m{\sseq{\Ttens}{m}} tensors are listed in \m{\Fcal} is also arbitrary. 
In particular, we cannot guarantee that the order will not change 
if we permute the vertices of the underlying graph by \m{\sigma}. 
\item Each tensor \m{\Ttens_a} is sensitive to the order in which the 
\m{\cbrN{i^{a}_1,\ldots,i^{a}_{p_a}}} vertices are listed, and this can also change with \m{\sigma}. 
In particular, if \m{\Ttens} is a vector type quantity (w.r.t. permutations of the subgraph), then 
under a local permutation 
\sm{(i^{a}_1,\ldots,i^{a}_{p_a})\stackrel{\tau}{\longmapsto}(i^{a}_{\tau(1)},\ldots,i^{a}_{\tau(p)})} 
it will change 
as \sm{\Ttens^{(\tau)}_{j}\<=\Ttens_{\tau^{-1}(j)}}. 
If it is a matrix type quantity then it willl change as 
\m{\Ttens^{(\tau)}_{j,j'}=\Ttens_{\tau^{-1}(j),\tau^{-1}(j')}}. 
More generally, if \m{\Ttens} is a \m{k}'th order tensor, then
\sm{\Ttens^{(\tau)}_{\seq{i}{k}}=\Ttens_{\tau^{-1}(i_1),\ldots\tau^{-1}(i_k)}}.
\end{compactenum}
Remarkably, despite all these degrees of freedom, it is still possible to define covariance and 
equivariance in higher order subgraph neural networks in a meaningful way. 

\begin{defn}\label{def: covariant HOSNN}\textbf{(Permutation covriant layer)}
Let \m{\Fcal\<=\cbrN{\sseq{\Ttens}{m}}} be the output of a layer in a higher order subgraph 
neural network induced by an invariant subgraph selection 
policy \m{\psi} and \m{\Fcal'\<=\cbrN{\sseq{\Ttens'}{m}}} be the output of the same layer after  
permuting the vertices of the underlying graph \m{\Gcal} by a permutation \m{\sigma}.  
By the invariance of \m{\psi}, for each \m{\Ttens_a} in \m{\Fcal} there is a corresponding \m{\Ttens'_{a'}} 
in \m{\Fcal'} such that \m{\cbrN{\sigma(i^{a}_1),\ldots,\sigma(i^{a}_{p_a})}} and 
\sm{\cbrN{{i'}^{a'}_1,\ldots,{i'}^{a'}_{p_{a'}}}} are equal as sets (intuitively, \m{\Ttens_a} 
and \m{\Ttens'_{a'}} correspond to the same underlying subgraph despite the relabeling by \m{\sigma}). 
In particular, there is a permutation \m{\tau\tin\Sbb_{p_a}} that elementwise aligns these two 
sets of vertex indices in the sense that \sm{i'^{a'}_{j}=\sigma(i^a_{\tau^{-1}(j)})}. \m{\Fcal} is said to be a 
permutation covariant layer if 
\begin{equation}
[\Ttens'_{a'}]_{\seq{j}{k}}=[\Ttens_a]_{\tau^{-1}(j_1),\ldots\tau^{-1}(j_k)}
\end{equation}  
for any permutation \m{\sigma\tin\Sn}.
\end{defn}

As usual, a permutation \m{equivariant} map is defined as one which preserves this covariance property. 

\begin{defn}\label{def: equivariant HOSNN}\textbf{(Permutation equivariant higher order subgraph neural network)}
A subgraph layer \m{\phi\colon\Fcal^{\textrm{in}}\mapsto \Fcal^{\textrm{out}}}, 
or more generally \m{\phi\colon(\Fcal^{\textrm{in}}_1,\ldots,\Fcal^{\textrm{in}}_r) \mapsto \Fcal^{\textrm{out}}}
 is said to be \df{permutation equivariant} if 
whenever \sm{\Fcal^{\textrm{in}}_1,\ldots,\Fcal^{\textrm{in}}_r} are covariant, 
\sm{\Fcal^{\textrm{out}}} is covariant as well. 
The entire network is equivariant if all the maps connecting its different layers are equivariant. 
\end{defn}

While these definitions appear quite technical, they are necessary for capturing the interaction between the 
effect of permutations at the subgraph selection level and the level of 
individual subgraphs. In the special case of \m{\psi} just picking out the individual vertices of \m{\Gcal} 
(as trivial subgraphs consisting of a single vertex) and the activations being scalars, the 
definitions reduce to the classical case where \m{\Fcal} can be represented as an \m{n\times c} matrix,  
where \m{n} is the number of vertices and \m{c} the number of channels. 
This base case is what one typically uses in the input layer and readout layer of an equivariant GNN. 
The purpose of the next section is to 
introduce a mathematical formalism that makes implementing the intermediate, 
higher order layers as straightforward as possible.


%% file: fig-subgraphmp.tex
\begin{figure}[t]
\vspace{-6pt}
\includepicc{.40}{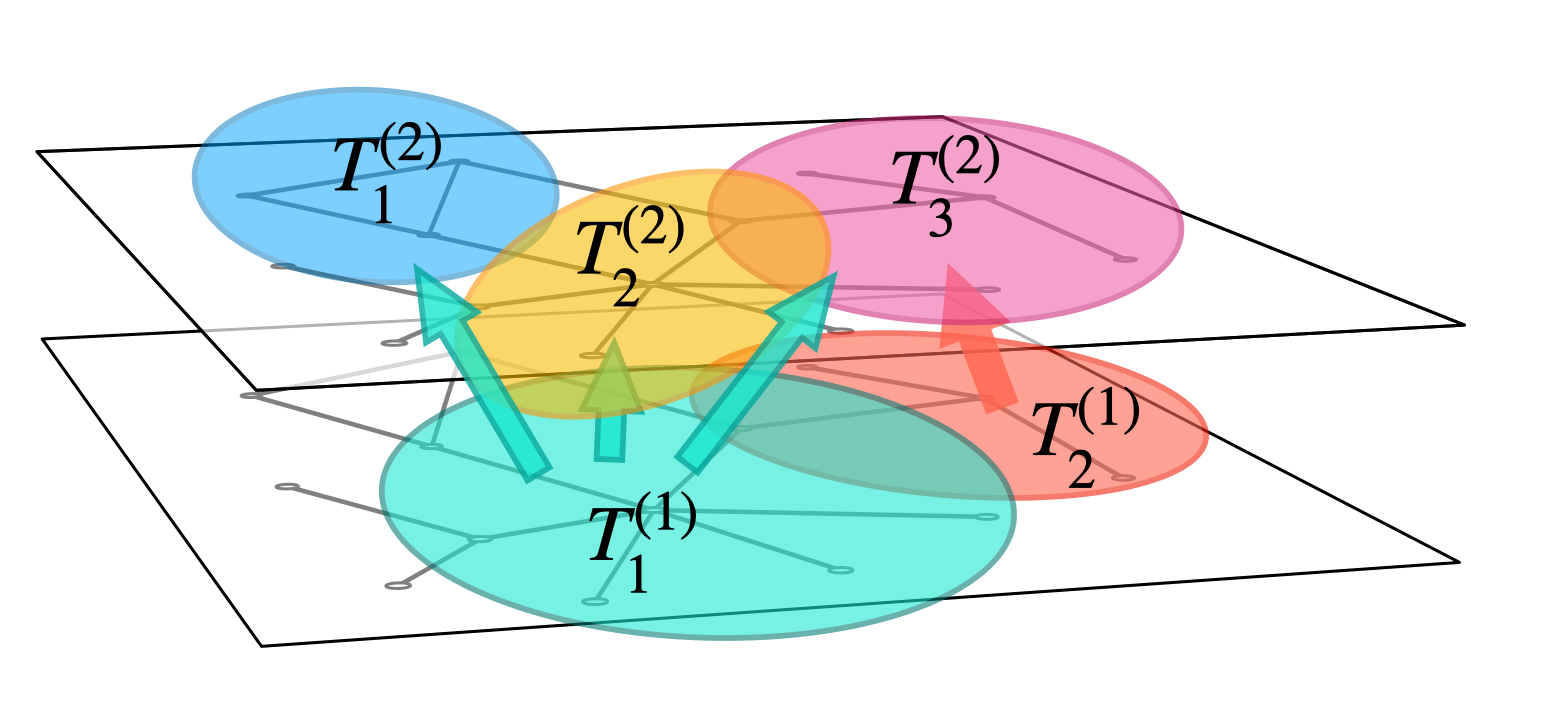}
\vspace{-6pt}
\caption{\label{fig: subgraphmp}
A subgraph neural network must be equivariant to two different ways 
that permutations act on it: changing the set of vertices 
assigned to a given subgraph, and reordering the vertices of the subgraph internally. 
This is especially important when the subgraph neurons produce matrix/tensor 
valued outputs indexed by the vertices of the subgraph itself. 
The \m{P}-tensors formalism allows us to handle this situation in a simple way, 
defining the most general form of equivariant linear messages between such 
tensors, without making reference to the global ordering.} 
\vspace{-5pt}
\end{figure}

%% file: ptensors.tex
\section{\m{P}-TENSORS}\label{sec: ptensors}

The mathematical device that we introduce to derive the rules of higher order message passing 
are a type of object that we call \emph{P-tensors}. 
To define \m{P}-tensors first we need to define a finite or countably infinite set \m{\Ucal} of base objects called 
\emph{atoms} that permutations act on. In the case of graph neural networks the atoms are  
just the vertices. 
However, the \m{P}-tensor formalism is also applicable to other permutation equivariant learning 
scenarios, such as relational learning, in which case  
\m{\Ucal} might for example be a set of individuals or the words in 
a given language. 

A given \m{P}-tensor \m{T} is defined relative to an ordered subset \m{\Dcal=(x_1,\ldots,x_d)} of atoms called its 
\df{reference domain}, which, in the case of subgraph neurons, 
is just the vertex set of the given subgraph. 
Reordering the reference domain by a permutation \m{\tau\tin\Sbb_d} changes it to 
\begin{equation}\label{eq: Dtau}
\Dcal'=\tau\circ\Dcal=(x_{\tau^{-1}(1)},\ldots,x_{\tau^{-1}(1)}).
\end{equation}
The defining property of \m{P}-tensors is how they transform 
under this action. 

\begin{defn}[\textbf{\m{P}-tensors}]
Let \m{\Ucal} be a finite or countably infinite set of atoms and \m{\Dcal\<=\brN{x_1,\ldots, x_d}} 
an ordered subset of \m{\Ucal}. 
We say that a \m{k}'th order tensor \m{T\tin\RR^{d\times d\times\ldots\times d}} is a \m{k}'th 
order \emph{permutation covariant tensor} (or \m{P}-tensor 
for short) with reference domain \m{\Dcal}  if under reordering \m{\Dcal} as in \rf{eq: Dtau} 
it transforms to 
\begin{equation}\label{eq: T-action}
[\tau\circ T]_{\seq{i}{k}}=T_{\tau^{-1}(i_1),\ldots\tau^{-1}(i_k)}.
\end{equation}
\end{defn}

The neurons in modern neural networks typically have many channels, so we also allow  
\m{P}-tensors to have a channel dimension. Thus, a \m{k}'th order \m{P}-tensor 
can actually be a \m{k\<+1}'th order tensor \sm{T\tin\RR^{d\times d\times\ldots\times d\times C}}. 
The channel dimension is not affected by permutations. 
We will sometimes also write \m{(T,\Dcal)} to denote a \m{P}-tensor with reference domain \m{\Dcal}. 

To derive the rules of equivariant message passing from one \Ptensor{} \m{T_1} to another 
\Ptensor{} \m{T_2}, we need to consider the three cases when their respective reference domains 
\m{\Dcal_1} and \m{\Dcal_2} in the unordered sense are (a) the same (b) partially overlap (c) are disjoint. 
The advantage of the \Ptensors{} formalism compared to the previous section  	 
is that in each of these cases we need only consider the action of permutations that are \emph{internal} to 
\m{\Dcal_1} and \m{\Dcal_2}. In particular, we have the following definition. 


\begin{defn}[\textbf{Permutation equivariant maps between \m{P}-tensors}]\label{def: Ptensor-equi} 
Let \m{\Dcal_1} and \m{\Dcal_2} be two fixed reference domains,  
\m{\wbar{\Dcal_1}} be any reordering of \m{\Dcal_1} and \m{\wbar{\Dcal_2}} any reordering a \m{\Dcal_2}. 
Consider a family of linear maps 
\[\phi_{\wbar{\Dcal_1},\wbar{\Dcal_2}}\colon (T_1,\wbar{\Dcal_1})\mapsto(T_2,\wbar{\Dcal_2}).\]
We say that this is a permutation equivariant family of linear maps between \m{P}-tensors if 
\[\phi_{\tau_1\circ \Dcal_1,\,\tau_2\circ \Dcal_2}(\tau_1\circ T_1)=\tau_2\circ (\phi_{\Dcal_1,\Dcal_2}(T_1))\]
for any \m{P}-tensor \m{T_1} with reference domain \m{\Dcal_1} and 
any pair of permutations \m{\tau_1\tin\Sbb_{\abs{\Dcal_1}}} and \m{\tau_2\tin\Sbb_{\abs{\Dcal_2}}}. 
\end{defn}

In the next section we will see that 
such families of equivariant maps can be defined in a relatively straightforward manner 
by transforming both the source and destination \m{P}-tensors  
to a canonical position, where the shared atoms between their reference domains occupy the first 
\m{d^\cap=\abs{\Dcal_1\cap\Dcal_2}} positions. 

In addition to equivariance to local permutations, we also need to consider global 
relabelings \m{\sigma\colon \Ucal\to\Ucal} of the entire universe of atoms. 
The effect of such a relabeling is to map \m{\Dcal=(x_1,\ldots,x_d)} to 
\m{\sigma\bullet \Dcal=(\sigma(x_1),\ldots,\sigma(x_d))}. 

\begin{defn}[\textbf{Relabeling invariant maps between \m{P}-tensors}]\label{def: Ptensor-relabeling}
A family of linear maps between \m{P}-tensors 
\[\phi_{\Dcal_1,\Dcal_2}\colon (T_1,\Dcal_1)\mapsto (T_2,\Dcal_2)\]
is said to be invariant to global relabelings if 
\[\phi_{\sigma\bullet\Dcal_1,\sigma\bullet\Dcal_2}=\phi_{\Dcal_1,\Dcal_2}\]
for any permutation \m{\sigma} of the universe \m{\Ucal} of atoms. 
\end{defn}

\ignore{
\begin{defn}[\textbf{Equivariant map between \m{P}-tensors}]\label{def: Ptensor-equi} 
Let \m{\Dcal_1} be the reference domain  of \m{T_1} and \m{\Dcal_2} be the reference domain of \m{T_2}. 
We say that a family of linear maps \m{\phi_{\Dcal_1,\Dcal_2}\colon T_1\to T_2} is permutation equivariant if 
\[\phi_{\tau_1\circ \Dcal_1,\,\tau_2\circ \Dcal_2}(\tau_1\circ T_1)=\tau_2\circ (\phi_{\Dcal_1,\Dcal_2}(T_1))\]
for any pair of permutations \m{\tau_1\tin\Sbb_{\abs{\Dcal_1}}} and \m{\tau_2\tin\Sbb_{\abs{\Dcal_2}}}. 
\end{defn}
}
 
A key result of our paper, proved in the Appendix,  
is the following theorem, showing that 
the above two constraints are exactly what is needed to build equivariant higher order MPNNs out of \Ptensors{}. 

\begin{theorem}\label{thm: equivariance}
Any higher order MPNN in which 
\vspace{-6pt}
\begin{compactenum}[~(a)]
\item the subgraphs in each layer are selected using an invariant 
subgraph selection rule; 
\item the output of each subgraph neuron is a P-tensor; 
\item the messages sent from the \m{P}-tensors in each layer \m{\Fcal^{\textrm{in}}} 
to the \m{P}-tensors in following layer \m{\Fcal^{\textrm{out}}} are linear and satisfy the conditions of 
Definitions \ref{def: Ptensor-equi} and \ref{def: Ptensor-relabeling} 
\end{compactenum}
\vspace{-6pt}
is a permutation equivariant 
MPNN in the sense of Definition \ref{def: equivariant HOSNN}.
\end{theorem}

\ignore{
\begin{theorem}\label{thm: equivariance}
Any higher order MPNN in which (a) the subgraphs in each layer are selected using an invariant 
subgraph selection rule; (b) the output of each subgraph neuron is a P-tensor; (c) 
each map \sm{\wbar \phi} connecting some layer \m{\Fcal^{\textrm{in}}} 
to some other layer \m{\Fcal^{\textrm{out}}} is linear, and such that the  ``message'' \m{\phi}  
sent from each \m{T^{\textrm{in}}_{i}} in \m{\Fcal^{\textrm{in}}} to each 
\m{T^{\textrm{out}}_{j}} in \m{\Fcal^{\textrm{out}}} only 
depends on the relative ordering of their reference domains; is a permutation equivariant 
MPNN in the sense of Definition \ref{def: equivariant HOSNN} if and only if 
\m{\phi} satisfies the condition of Definition \ref{def: Ptensor-equi}. 
\end{theorem}
}
\ignore{
\begin{theorem}
Let \m{\Ncal} be a higher order MPNN in which the subgraphs in each layer are selected using an 
invariant selection policy. Assume that layer 
\m{\Fcal^{\textrm{out}}=\cbrN{F_1^{\textrm{out}_1},\ldots,F_1^{\textrm{out}_{m_1}}}} 
is computed from some other layer 
\m{\Fcal^{\textrm{in}}=\cbrN{F_1^{\textrm{in}_1},\ldots,F_1^{\textrm{in}_{m_2}}}} 
via a linear map such \m{}that  
\end{theorem}
}

\ignore{
While seemingly abstract, these definitions capture exactly how higher order quantities related to different 
but potentially overlapping sets of objects are expected to behave under permutations, 
in particular, how the activations of HO-MPNNs change under permutations of the 
vertices. 
In line with much of the literature on different types of generalized convolutional neural networks, 
in the following we focus on deriving the most general \emph{linear} equivariant operations 
between P-tensors.  
We do this in two steps: first we consider the case that \sm{\Dcal_1} and \m{\Dcal_2} are equal 
as sets, and then the more  general case where \m{\Dcal_1} and \m{\Dcal_2} overlap only partially. 
}

\subsection{Message passing between \m{P}-tensors with the same reference domain}\label{subsec: Maron}

If \m{\Dcal_1=(\sseq{x}{d})} and \m{\Dcal_2=(\sseq{x'}{d})} are equal \emph{as sets}, 
they can  only differ by a permutation \m{\mu} mapping \m{x'_1\<=x_{\mu(1)}}, \m{x'_2\<=x_{\mu(2)}}, and so on. 
In any kind of permutation equivariant learning scenario such as graph neural networks, 
it is always possible to determine which labels refer to the same object, 
so, without loss of generality, we can 
assume that the elements of \m{\Dcal_2} 
have been rearranged so that \m{x_1'\<=x_1},\ldots\m{x_d'\<=x_d}. 
This reduces the problem of satisfying Definition \ref{def: Ptensor-equi} 
to that of  ``ordinary'' permutation equivariant vector, matrix, 
etc., valued neural network layers, which, by now, is a well studied subject.
\input{tbl-Maron-2-2}

In the first order case, \Ptensors{} are simply vectors, so deriving the rules of 
permutation equivariant message passing reduces to finding the space of linear maps 
\m{\phi\colon \RR^d\to\RR^d} 
satisfying 
\[\phi([\v]_{\tau^{-1}(i)})=[\phi(\v)]_{\tau^{-1}(i)}\]
for any \m{\v\tin\RR^d} and any \m{\tau\tin\Sbb_d}. 
The seminal Deep Sets paper \cite{zaheer2017deep} proved that in this case \m{\phi} can  
have at most two (learnable) parameters \m{\lambda_1} and \m{\lambda_2}, and must be of the form  
\[\phi(\v)=\lambda_1 \v+\lambda_2 \V 1\ts \V 1^\top\! \v.\]
In the more general case, \m{\phi} maps a \m{k_1}'th 
order tensor \m{\Ttin} to a \m{k_2}'th order tensor \m{\Tout}, 
both transforming under permutations as in \rf{eq: T-action}, 
leading to the equivariance condition    
\begin{equation}\label{eq: Maron equivariance} 
[\phi(\Ttin)]_{\tau^{-1}(i_1),\ldots,\tau^{-1}(i_{k_2})}=
\phi(\Ttin_{\tau^{-1}(i_1),\ldots,\tau^{-1}(i_{k_1})}).
\end{equation}
The characterization of the space of equivariant maps for this case was given 
in a similarly influential paper \citep{maron2018}. 

\begin{prop}[Maron et al.]\label{prop: Maron} 
The space of linear maps \sm{\phi\colon \RR^{d^{k_1}}\to\RR^{d^{k_2}}} that is equivariant to permutations 
\m{\tau\tin\Sbb_{d}} in the sense of \rf{eq: Maron equivariance} is spanned by a basis 
indexed by the partitions of the set \m{\cbrN{\oneton{k_1\<+k_2}}}. 
\end{prop}

Since the number of partitions of \m{\cbrN{\oneton{m}}} is given by the so-called \emph{Bell number} \m{B(m)}, 
according to this result, the number of learnable parameters in such a map is \m{B(k_1\<+k_2)}. 
To be specific, the equivariant linear map corresponding to a given partition \m{\Pcal} can be 
written as a composition of three operations:  
(1) \emph{summing} over specific dimensions or diagonals of \m{\Ttin}; 
(2) \emph{transferring} this result to the output tensor by identifying some combinations of input indicies 
with output indices; (3) \emph{broadcasting} the result along certain dimensions or diagonals of the output 
tensor \m{\Fout}. These three operations correspond to the three different types of parts that can appear in 
\m{\Pcal}: 
those parts that only involve the second \m{k_2} numbers, those that involve a mixture of the 
first \m{k_1} and second \m{k_2}, and those that only involve the first \m{k_1}. 
We shall say that \m{\Pcal} is of type \m{(p_1,p_2,p_3)} if it has \m{p_1} parts of the first category, 
\m{p_2} of the second and \m{p_3} of the third. 
In all three categories, a part \m{\cbrN{j_1,\ldots,j_\ell}} 
appearing in \m{\Pcal} implies that the corresponding indices  
are tied together. 
The way to distinguish between input and output indices is that if \m{1\leq j_q\leq k_1} then 
it refers to the \m{j_q}'th index of \m{\Fout}, whereas if \m{k_1\<+1\leq j_q\leq k_1+k_2}, 
then it refers to the \m{j_q-k_1} index of \m{\Fin}. 

As an example, in the \m{k_1\<=k_2\<=3} case, the partition \m{\Pcal=\cbrN{\cbrN{1,3},\cbrN{2,5,6},\cbrN{4}}} 
corresponds to 
(a) summing \m{\Ttin} along its first dimension (corresponding to \m{\cbrN{4}}) 
(b) transferring the diagonal along the second and third dimensions of \m{\Ttin} 
to the second dimension of \m{\Fout} (corresponding to \m{\cbrN{2,5,6}}), 
(c) broadcasting the result along the diagonal of the first and third dimensions 
(corresponding to \m{\cbrN{1,3}}). 
Explicitly, this gives the equivariant map 
\begin{equation}\label{eq: M-example1}
\Fout_{a,b,a}=\sum_{c}\Ttin_{c,b,b}.
\end{equation}
Since \m{B(6)\<=203}, listing all other possible maps for \m{k_1\<=k_2\<=3} would be very laborious.  
In Table \ref{tbl: Maron-2-2} we list the 
possible equivariant maps for the \m{k_1\<=k_2\<=2} case.

The presence of multiple channels enriches this picture only to the extent that each input channel 
can be linearly mixed with each output channel. 
For example, in the case of \m{C_{\textrm{in}}} channels in \m{\Fin} and \m{C_{\textrm{out}}} channels in 
\m{\Fout}, 
\rf{eq: M-example1} becomes 
\[\Fout_{a,b,a,\alpha}=\sum_{\beta} W_{\alpha,\beta} \sum_{c}\Fin_{c,b,b,\beta}\]
for some (learnable) weight matrix \m{W\tin\RR^{C_{\textrm{out}}\times C_{\textrm{in}}}}. 
This type of linear mixing across channels can be separated 
from the equivariant message passing operation itself, 
which consists of applying \rf{eq: M-example1} to each channel separately.

\subsection{Message passing between P-tensors with different reference domains}
\input{fig-cutout}

Our second main result is the following theorem, which 
generalizes Proposition \ref{prop: Maron} 
to when the reference domains of the input and output tensors only partially overlap. 

\begin{theorem}\label{thm: main theorem}
Let \m{T_1} and \m{T_2} be two P-tensors with reference domains \m{\Dcal_1} and \m{\Dcal_2} such that 
\m{\abs{\Dcal_1\cap\Dcal_2}\<\geq 2} and \m{\Dcal_1\not\subseteq \Dcal_2} and \m{\Dcal_2\not\subseteq \Dcal_1}. 
Then for each partition \m{\Pcal} of \m{\cbrN{1,\ldots,{k_1\<+k_2}}} of type \m{(p_1,p_2,p_3)} there are 
\m{2^{p_1+p_3}} linearly independent permutation equivariant linear maps \m{\phi\colon T_1\mapsto T_2}.  
\end{theorem}
\noindent

To derive the form of the actual maps, without loss of generality, we assume that \m{\Dcal_1} and \m{\Dcal_2} 
have been reordered in such a way that the \m{\dcap=\absN{\Dcal_1\cap\Dcal_2}} 
atoms that they have in common occupy the first \m{\dcap} 
positions in both, and are listed in the same order. 
An easy generalization of our previous results is to associate to each partition the same 
map as before, except we now only transfer information from the subtensor 
of \m{\Fin} cut out by the common atoms to the subtensor 
of \m{\Fout} cut out by the same (Figure \ref{fig: cutout} left).  
For example, the counterpart of \rf{eq: M-example1} would be 
\begin{equation*}
\Fout_{a,b,a}=
\begin{cases} 
\sum_{c=1}^{\dcap}\Fin_{c,b,b} & a,b\leq \dcap\\
0 & \Totherwise.
\end{cases}
\end{equation*}
This, however, would only give us \m{B(k_1\<+k_2)} equivariant maps, like in the previous section. 

The additional factor of \m{2^{p_1+p_3}} comes from the fact that for any partition that has parts 
purely involving indices of \m{\Fin} or \m{\Fout}, each of the corresponding operations can extend 
across either just the common atoms, or all atoms of the given tensor. 
For our running example \m{\Pcal=\cbrN{\cbrN{1,3},\cbrN{2,5,6},\cbrN{4}}}, 
we have the three additional equivariant maps
\begin{eqnarray*}
\Fout_{a,b,a}=
\begin{cases} 
\sum_{c=1}^{\dcap}\Fin_{c,b,b} & b\leq \dcap\\
0 & \Totherwise,
\end{cases} 
&& a\tin\cbrN{1,\ldots, d_2}\\ 
\Fout_{a,b,a}=
\begin{cases} 
\sum_{c=1}^{d_1}\Fin_{c,b,b} & a,b\leq \dcap\\
0 & \Totherwise,
\end{cases}
&& a\tin\cbrN{1,\ldots, \dcap}\\ 
\Fout_{a,b,a}=
\begin{cases} 
\sum_{c=1}^{d_1}\Fin_{c,b,b} & b\leq \dcap\\
0 & \Totherwise.
\end{cases}
&& a\tin\cbrN{1,\ldots, d_2}.\\
\end{eqnarray*}
Table \ref{tbl: Bell2} 
contrasts the number of possible equivariant maps in this case to 
the number of maps described in the previous section. 
As before, in the presence of multiple channels, 
each of the above maps can also involve mixing the different channels by a learnable weight matrix. 
\input{tbl-Bell2.tex} 

\ignore{
The actual form of the equivariant linear maps differ from those in the previous section in two 
respects. First, the transfer indices only extend over the set of \emph{common} elements \m{\Dcal_1\cap \Dcal_2}. 
Effectively this means that we only transfer information between the subtensors of \m{T_1} and \m{T_2} 
carved out by the common index values. Second, for each summation and broadcast index we consider two 
variations: the case when the summation/broadcast only takes place over the common index values and when 
it involves all values (hence the \m{2^{p_1+p_3}} factor in the Theorem). 

Once again, this is best demonstrated on a specific example. Without loss of generality, we assume that 
the ordering of \m{\Dcal_1} and \m{\Dcal_2} has been rearranged so that \m{k} common indices appear first 
and are matched between the two tensors. For the same example as before of maps between third order 
tensors corresponding to the \m{\Pcal=\cbrN{\cbrN{1,3},\cbrN{2,5,6},\cbrN{4}}} 
partition we now have four different maps:
\begin{eqnarray*}
&\Fout_{a,b,a}=\sum_{c=1}^{\abs{\Dcal_1\cap\Dcal_2}}\Fin_{c,b,b}\\
&\Fout_{a,b,a}=\sum_{c=1}^{\abs{\Dcal_1}}\Fin_{c,b,b}\\
&\Fout_{a,b,a}=
	\sum_{c=1}^{\abs{\Dcal_1\cap\Dcal_2}}\Fin_{c,b,b}\\
\end{eqnarray*}
}\ignore{ 
if for each 

P-tensors are defined with reference to permutations acting on their own reference domain. 
Ultimately, however, we are interested in how to send messages between two P-tensors in a way that 
is equivariant to both their reference domains being permuted simultaneously, in a coordinated way. 
Defining what we mean by this precisely brings us back to the notion of two-level equivariance 
mentioned at the end of the previous section. 
To describe equivariant message passing from one P-tensor \m{T_1} to a second P-tensor \m{T_2} 
we need to consider \emph{global} 
permutations \m{\sigma} 
of \m{\Ucal} that fix their reference domains \m{\Dcal_1} and \m{\Dcal_2} as \emph{sets}, 
but may change the relative ordering of \m{\Dcal_1} w.r.t.~\m{\Dcal_2}, denoted by \m{\tau}.  
To make these notions precise, we need the following definitions. 

\begin{defn}[\textbf{Action of global permutations}] 
Let \m{\Ucal=\cbrN{\seqi{x}}} be a finite or countably infinite set of atoms and \m{\sigma} be a permutation of the elements 
of ~\m{\Ucal}~ so that \m{x'_i\<=x_{\sigma^{-1}(i)}}. 
Given an ordered subset \m{\Dcal\<=(x_{i_1},\ldots, x_{i_d})} of~ \m{\Ucal},  
we define \sm{\sigma(\Dcal)\<=(x_{\sigma^{-1}(i_1)},\ldots, x_{\sigma^{-1}(i_d)})}. 
\end{defn}

\begin{defn}[\textbf{Fixed sets and restricted permutations}]
We say that \m{\sigma} \df{fixes} \m{\Dcal} (as a set) if \m{\Dcal} and \m{\sigma(\Dcal)} 
are equal \emph{as sets} (denoted \sm{\Dcal\seteq{}\sigma(\Dcal)}), i.e., if 
\begin{equation*}
\sigma(\Dcal):=
\brN{x_{\sigma^{-1}(i_1)},\,x_{\sigma^{-1}(i_2)},\,\ldots\,,x_{\sigma^{-1}(i_d)}}=
\brN{x_{i_{\tau^{-1}(1)}},\ldots, x_{i_{\tau^{-1}(d)}}}
\end{equation*}
for some permutation \m{\tau\tin\Sbb_{d}}. 
We say that \m{\tau} is the \df{restriction} of \m{\sigma} to \m{\Dcal} and 
denote it \m{\tau\<=\sigma\dn_{\Dcal}}. 
\end{defn}
}

\ignore{
\begin{defn}[\textbf{Fixed set}]
Let \m{\Ucal=\cbrN{\seqi{x}}} be a finite or countably infinite set of atoms and \m{\sigma} be a permutation of the elements 
of ~\m{\Ucal}~ so that \m{x'_i\<=x_{\sigma^{-1}(i)}}. 
Given an ordered subset \m{\Dcal\<=(x_{i_1},\ldots, x_{i_d})} of~ \m{\Ucal},  
we define \m{\sigma(\Dcal)\<=(x_{\sigma^{-1}(i_1)},\ldots, x_{\sigma^{-1}(i_d)})}. 
We say that \m{\sigma} \df{fixes} \m{\Dcal} \df{as a set} if 
if this action permutes the internal ordering of the atoms in \m{\Dcal}, but does not 
change which atoms fall in \m{\Dcal} in the first place, i.e., if
\begin{mequation}
\cbrN{x_{\sigma^{-1}(i_1)},\,x_{\sigma^{-1}(i_2)},\,\ldots\,,x_{\sigma^{-1}(i_d)}}=
\cbrN{x_{i_1},\ldots, x_{i_d}}.
\end{mequation}
\end{defn}
}

\ignore{
\begin{defn}[\textbf{Restricted permutation}]
Let \m{\sigma} be a permutation of \m{\Ucal} that fixes \m{\Dcal\<=\brN{x_{i_1},\ldots, x_{i_d}}} as a set. 
We define the \df{restricted permutation} \m{\tau=\sigma\dn_{\Dcal}\tin\Sbb_{\abs{\Dcal}}} 
as the permutation satisfying 
\[
\sigma^{-1}(i_j)=i_{\tau^{-1}(j)}
\qquad j\tin\cbrN{\oneton{d}}.\]
\end{defn}
}
\ignore{
Clearly, \m{\tau\<=\sigma\dn_{\Dcal}}, when it exists, must satisfy \sm{\sigma(i_j)\<=i_{\tau(j)}} 
for all \m{j}.  
Moreover, if the original \m{\Dcal} is written as \sm{\Dcal=\br{\sseq{\wbar{x}}{d}}}, 
whenever \m{\Dcal} is fixed by \m{\sigma}, its permuted version 
can be written as just 
\m{\sigma(\Dcal)=\brN{\wbar{x}_{\tau^{-1}(1)}, \wbar{x}_{\tau^{-1}(2)}, \ldots, \wbar{x}_{\tau^{-1}(d)}}}.  
We are now in a position to define what it means for a map \m{\phi\colon T_1\mapsto T_2} between two P-tensors 
to be equivariant. 
}

%% file: tbl-maron-2-2.tex
\begin{table}[t]
\begin{equation*}
\small{
\begin{array}{|c|c|}
\hline
\Pcal& \phi\\
\hline
\cbrN{\cbrN{1},\cbrN{2},\cbrN{3},\cbrN{4}} & \Fout_{a,b}=\sum_{c,d}\Fin_{c,d}\\ 
\cbrN{\cbrN{1},\cbrN{2},\cbrN{3,4}} & \Fout_{a,b}=\sum_{c}\Fin_{c,c}\\ 
\cbrN{\cbrN{1},\cbrN{2,4},\cbrN{3}} & \Fout_{a,b}=\sum_{c}\Fin_{c,b}\\ 
\cbrN{\cbrN{1},\cbrN{2,3},\cbrN{4}} & \Fout_{a,b}=\sum_{c}\Fin_{b,c}\\ 
\cbrN{\cbrN{2},\cbrN{1,4},\cbrN{3}} & \Fout_{b,a}=\sum_{c}\Fin_{c,b}\\ 
\cbrN{\cbrN{2},\cbrN{1,3},\cbrN{4}} & \Fout_{b,a}=\sum_{c}\Fin_{b,c}\\ 
\cbrN{\cbrN{1,2},\cbrN{3},\cbrN{4}} & \Fout_{a,a}=\sum_{b,c}\Fin_{b,c}\\ 
\cbrN{\cbrN{1},\cbrN{2,3,4}} & \Fout_{a,b}=\Fin_{b,b}\\ 
\cbrN{\cbrN{2},\cbrN{1,3,4}} & \Fout_{b,a}=\Fin_{b,b}\\ 
\cbrN{\cbrN{1,2,3},\cbrN{4}} & \Fout_{a,a}=\sum_b \Fin_{a,b}\\ 
\cbrN{\cbrN{1,2,4},\cbrN{3}} & \Fout_{a,a}=\sum_b \Fin_{b,a}\\ 
\cbrN{\cbrN{1,2},\cbrN{3,4}} & \Fout_{a,a}=\sum_{c}\Fin_{c,c}\\ 
\cbrN{\cbrN{1,3},\cbrN{2,4}} & \Fout_{a,b}=\Fin_{a,b}\\ 
\cbrN{\cbrN{1,4},\cbrN{2,3}} & \Fout_{a,b}=\Fin_{b,a}\\ 
\cbrN{\cbrN{1,2,3,4}}& \Fout_{a,a}=\Fin_{a,a}\\
\hline
\end{array}
}
\end{equation*}
\vspace{-10pt}
\caption{\label{tbl: Maron-2-2}
The \m{B(4)\<=15} possible partitions of the set \m{\cbrN{1,2,3,4}} and 
the corresponding permutation equivariant linear maps \m{\phi:\RR^{k\times k}\to \RR^{k\times k}} 
as derived by Maron et al. \cite{maron2018}.}
\end{table}

%% file: fig-cutout.tex
\begin{figure}[t]
\vspace{-6pt}
\includepicc{.32}{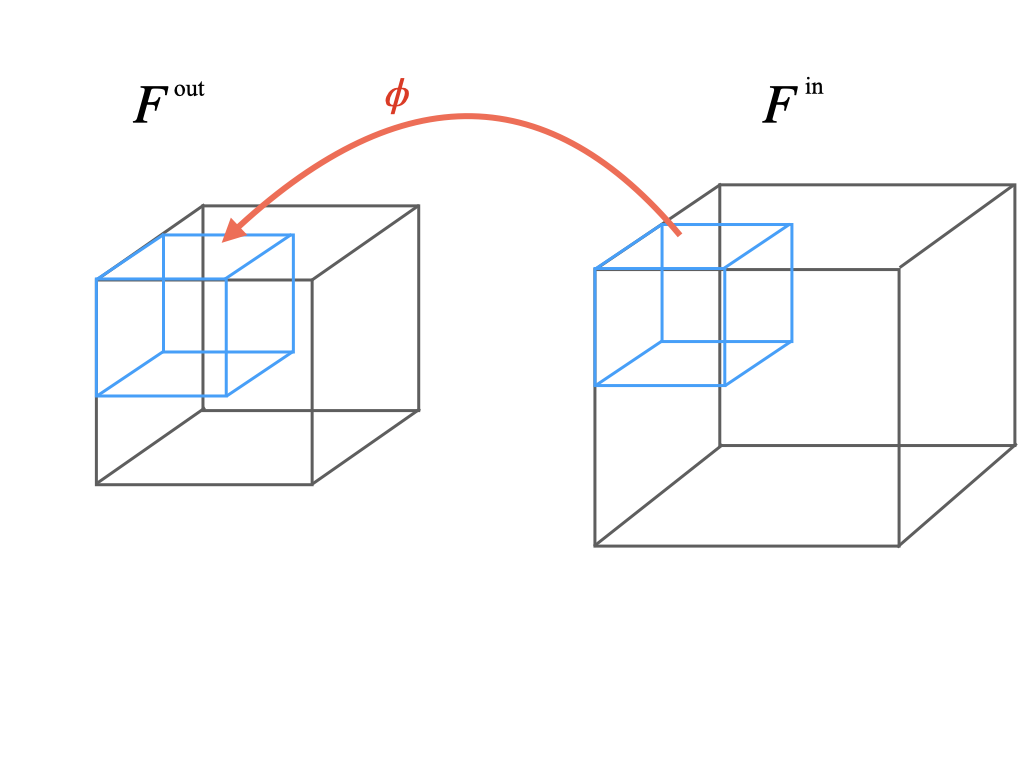}
\ignore{
\begin{minipage}{.49\textwidth}
\begin{equation*}
\small 
\begin{array}{|c|c|c|}
\hline
(k_1,k_2)&\pbox{100pt}{\# of maps in\\ \m{\Dcal_1\seteq\Dcal_2} case} &\pbox{100pt}{\# of maps in\\ \m{\Dcal_1\not{\!\seteq}\Dcal_2} case} \\
\hline
(0,0)&   1 &   1 \\
(1,1)&   2 &  17 \\
(1,2)&   5 &  49 \\
(2,2)&  15 & 129 \\
(2,3)&  52 & 321 \\
(3,3)& 203 & 769 \\
\hline
\end{array}
\end{equation*}
\end{minipage}
}
\vspace{-35pt}
\caption{\label{fig: cutout}
Given two \m{P}-tensors \m{\Fin} and \m{\Fout} whose reference domains have \m{\dcap} atoms in common, 
without loss of generality we can rearrange the two tensors so that the indices corresponding to the common atoms 
appear first. Mapping the corresponding \m{\dcap\!\times \dcap\!\times \ldots \times \dcap} subtensor of \m{\Fin} to the 
analogous \m{\dcap\!\times \dcap\!\times \ldots\times \dcap} subtensor of \m{\Fout} with any of the \m{B(k_1\<+k_2)} 
linear maps described in Section \ref{subsec: Maron} is an equivariant operation. 
The additional equivariant operations correspond to similar maps except with the summations and 
broadcast operations extending over not just the overlapping part of the tensors but the 
entirety of \m{\Fin} or \m{\Fout}. 
}
\end{figure}

%% file: tbl-Bell2.tex
\begin{figure}[t]
\begin{equation*}
{\small 
\begin{array}{|c|c|c|}
\hline
(k_1,k_2)&\pbox{100pt}{\# of maps in\\ \m{\Dcal_1\<=\Dcal_2} case} &\pbox{100pt}{\# of maps in\\ \m{\Dcal_1\neq\Dcal_2} case} \\
\hline
(0,0)&   1 &   1 \\
(1,1)&   2 &  5 \\
(1,2)&   5 &  17 \\
(2,2)&  15 & 61 \\
(2,3)&  52 & 321 \\
(3,3)& 203 & 769 \\
\hline
\end{array}
}
\end{equation*}
\vspace{-10pt}
\caption{\label{tbl: Bell2} 
The number of independent equivariant linear maps from a \m{k_1}'th order \m{P}-tensor to a 
\m{k_2} 'th order \m{P}-tensor when the reference domains are the same vs.~when they overlap only partially. 
}
\end{figure}

%% file: gnns.tex
\section{\m{P}-TENSORS IN GRAPH NEURAL NETWORKS}

In this section we describe how the \m{P}-tensor formalism can be used to build expressive graph 
neural networks, and how various existing GNNs reduce to special cases. 

\subsection{Zeroth order message passing networks}
\newcommand{\chan}{c}

In classical message passing networks the reference domain of each neuron consists of just a single 
vertex. Talking about the transformation properties of an object with respect to permutations of  
a single atom is vacuous, so in this case all the activations are zeroth order \m{P}-tensors. 
The only type of equivariant message that a zeroth order vertex neuron \sm{T^{v_1}_\chan} can send 
to another vertex neuron \sm{T^{v_2}_\chan} is \m{T^{v_2}_\chan \leftarrow T^{v_1}_\chan} 
(here and in the following \m{\chan} stand for the channel index). 
Applying this operation to all the neighbors of a given vertex and adding a bias term plus mixing 
with a learnable matrix \m{W} leads exactly to the update rule \rf{eq: MPNN-update}. 
Thus, classical message passing networks just correspond to message passing between zeroth order 
\m{P}-tensors. 

\subsection{Edge networks}

One of the first extensions of the MPNN formalism was the introduction of networks that can pass messages 
not just from vertices to vertices but also from vertices to edges and edges to vertices 
\cite{gilmer2017neural}. 
In the \m{P}-tensor formalism a neuron corresponding to the edge \m{\wbar{v_1v_2}} has 
receptive domain \m{\Dcal=\brN{v_1,v_2}}. If the edge \Ptensor{} \m{T^{\wbar{v_1v_2}}} is zeroth order, 
then the rule for sending messages from the vertex \Ptensors{} \m{T^{v_1}} and \m{T^{v_2}} 
will essentially be the same as above. 

However, when \sm{T^{\wbar{v_1v_2}}} is a first order \m{P}-tensor 
we have two different possible equivariant maps: 
the ``concatenating map'' 
\begin{equation*}
T^{\wbar{v_1v_2}}_{i,\chan}\leftarrow T^{v_i}_{\chan}\qqquad\qquad  i\tin\cbrN{1,2},
\end{equation*}
and the ``averaging map''
\begin{equation*}
T^{\wbar{v_1v_2}}_{i,\chan}\leftarrow T^{v_1}_{\chan}+T^{v_2}_{\chan}\qqquad i\tin\cbrN{1,2}.
\end{equation*}
To maximize expressivity, the edge neuron would conctenate these two messages, effectively doubling 
the number of channels as we pass from the ``vertex layer'' neurons to the ``edge layer''.

Edge-to-vertex message passing similarly affords two distinct types of linear maps:
\[T^{v_i}_{\chan}\leftarrow T^{\wbar{v_1v_2}}_{i,\chan} \qquad \textrm{and}\qquad 
T^{v_i}_{\chan}\leftarrow T^{\wbar{v_1v_2}}_{1,\chan}+T^{\wbar{v_1v_2}}_{2,\chan}.
\]
Finally, for passing a message from a first order edge \sm{T^{(v_1,v_2)}} to 
another first order edge \sm{T^{(v_2,v_3)}} (with \m{v_1\<\neq v_3}) we have one linear 
map corresponding to the \sm{\cbrN{\cbrN{1,2}}} partition, 
\begin{equation*}
T^{\wbar{v_2v_3}}_{1,\chan}\leftarrow T^{\wbar{v_1 v_2}}_{2,\chan}
\end{equation*}
and three different maps corresponding to the \m{\cbrN{\cbrN{1},\cbrN{2}}} partition: 
\begin{align*}
&T^{\wbar{v_2v_3}}_{i,\chan}\leftarrow T^{\wbar{v_1v_2}}_{2,\chan} \qquad &i\tin\cbrN{1,2}\\
&T^{\wbar{v_2v_3}}_{1,\chan}\leftarrow T^{\wbar{v_1v_2}}_{1,\chan}+T^{\wbar{v_1v_2}}_{2,\chan}\\
&T^{\wbar{v_2v_3}}_{i,\chan}\leftarrow T^{\wbar{v_1v_2}}_{1,\chan}+T^{\wbar{v_1v_2}}_{2,\chan}\qquad &i\tin\cbrN{1,2}.
\end{align*}
While individually these operations are simple and could be derived by hand on a case-by-case basis, 
as the size of the reference domains (as well as the order of the tensors) 
increases, enumerating and separately implementing all possibilities in code becomes unwieldly. 

\subsection{Message passing between subgraphs}

The true power of the \Ptensors{} model manifests in message passing between larger subgraphs. 
Take for example the paradigmatic case from chemistry of two adjacent benzene rings 
(six-atom rings of carbon atoms) 
made up of vertices \m{\cbrN{\sseq{v}{6}}} and \m{\cbrN{v_1,v_2,v_7,v_8,v_9,v_{10}}} 
represented by first order \Ptensors{} \m{\Ts} and and \m{\Td}. 

The linear map corresponding to the \sm{\cbrN{\cbrN{1,2}}} partition,  
\begin{equation*}
\Td_{i,\chan}\leftarrow \begin{cases} \Ts_{i,\chan} & i\tin\cbrN{1,2}\\ 0 &\otherwise \end{cases}
\end{equation*}
transfers the rows of \m{\Ts} corresponding to the two shared carbon atoms to the corresponding rows of \m{\Td}. 
We have four maps corresponding to \m{\cbrN{\cbrN{1},\cbrN{2}}}: 
\begin{compactenum}[1.]
\item Transferring the sum of the rows of the shared carbons to the corresponding rows:
\begin{equation*}
\Td_{i,\chan}\leftarrow \begin{cases} \sum_{j=1}^2 \Ts_{j,\chan} & i\tin\cbrN{1,2}\\ 0 &\otherwise, \end{cases}
\end{equation*}
\item Transferring the sum of the rows of the shared carbons to all rows:
\vspace{-6pt}
\begin{equation*}
\Td_{i,\chan}\leftarrow \tsum_{j=1}^2 \Ts_{j,\chan},
\vspace{-4pt}
\end{equation*}
\item Transferring the sum of all rows to the rows of the shared carbons:
\vspace{-6pt}
\begin{equation*}
\Td_{i,\chan}\leftarrow \begin{cases} \sum_{j=1}^6 \Ts_{j,\chan} & i\tin\cbrN{1,2}\\ 0 &\otherwise, \end{cases}
\end{equation*}
\item Transferring the sum of all rows to all rows:
\vspace{-6pt}
\begin{equation*}
\Td_{i,\chan}\leftarrow \tsum_{j=1}^6 \Ts_{j,\chan}.\vspace{-4pt}
\end{equation*}
\end{compactenum}
If we concatenate the results of all of these maps, \m{\Td} will have five times as many channels as \m{\Ts}. 
\input{tbl-ZINC-HIV}

The second-order to second-order case is even more interesting, since in this case the \m{i,j} slice of 
\m{\Ts_{i,j,c}} and \m{\Td_{i,j,c}} can effectively represent interactions between the \m{i}'th and 
\m{j}'th atoms in the two rings. Space limitations prevent us from listing all 61 possible maps, 
but some examples are the following:
\begin{compactenum}[1.]
\item Transferring the sum of the interactions of shared carbon \m{i} with the other carbons in the ring 
to the corresponding slice:
\vspace{-4pt}
\begin{equation*}
\Td_{i,i,\chan}\leftarrow \begin{cases} \sum_{k=1}^6 \Ts_{i,k,\chan} & i\tin\cbrN{1,2}\\ 0 &\otherwise, \end{cases}
\end{equation*}
\item Transferring the sum of all interactions to the shared carbons:
\vspace{-8pt}
\begin{equation*}
\Td_{i,i,\chan}\leftarrow \begin{cases} \sum_{k=1}^6\sum_{\ell=1}^6 \Ts_{k,\ell,\chan} & i\tin\cbrN{1,2}\\ 0 &\otherwise, \end{cases}
\end{equation*}
\item Transferring the sum of all self-interactions to the shared carbons:
\vspace{-6pt}
\begin{equation*}
\Td_{i,i,\chan}\leftarrow \begin{cases} \sum_{k=1}^6\Ts_{k,k,\chan} & i\tin\cbrN{1,2}\\ 0 &\otherwise, \end{cases}
\end{equation*}
\end{compactenum}
Note that it is often the linear combinations of these maps, 
e.g., the second map above minus the third map, that have the most inutitive interpretations. 
Also note that many of these maps occur naturally in other subgraph, hypergraph and simplicial complex 
networks. The advantage of our formalism is in being to enumerate (and efficiently implement) 
all possible equivariant maps in a systematic way.


\ignore{
\subsection{Unite layers}

In a standard vertex-to-vertex message passing network each neuron in the first layer is only sensitive 
to the inputs at its neighboring vertices. A neuron in the second layer is sensitive to the inputs 
at its neighbors and neighbors of neighbors. In general the ``effective receptive field'' of the neuron 
in the \m{\ell}'th layer attached to vertex \m{v_i} consists of all the input neurons in the ball \m{B_\ell(v_i)} of 
radius \m{\ell} around \m{v_i}. To address the vertex identifiability issue mentioned in Section 
\ref{subsec: higher-order-MPNN}, \cite{maron2018} proposed an architecture in which correspondingly the 
reference domain of the P-tensor at vertex \m{v_i} in layer \m{\ell} is \m{B_\ell(v_i)}. 
In this case, the reference domain of each P-tensor is the \emph{union} of the reference domains 
of the P-tensors from which it aggregates information. For this reason, we call layers implementing 
this scheme \emph{unite layers}. 
While in \cite{maron2018} the rules for equivariant message passing were derived in an ad-hoc manner, 
the P-tensors formalism allows us to derive them automatically, as just a special case.

\subsection{Subgraph neurons}

Another recurring idea in graph neural networks is to explicitly identify certain subgraphs in \m{\Gcal} 
and assign specific message passing rules to them with individualized weights. 
One context in which this idea is particularly important is learning from the chemical structure of molecules, 
where the subgraphs can be chosen as functional groups. 
}

%% file: tbl-ZINC-HIV.tex
\begin{table*}[t]
        \centering
{\small
        \begin{tabular}{lcccc}\hline
                & ZINC-12K & ZINC-Full & OGBG-MOLHIV & TOX21\\
                                                                &MAE($\downarrow$)&MAE($\downarrow$)& ROC-AUC($\%\uparrow$) & ROC-AUC($\%\uparrow$)\\\hline
                RP-NGF \citep{murphy2019relational}             &         --                    & -- & -- & $0.79.4\pm 1.00$\\
                GCN \citep{kipf17}                              & $0.321\pm 0.009$              & -- &$76.07\pm0.97$ & --\\ 
                GIN \citep{xu2018powerful}                      & $0.408\pm 0.008$              & $0.088\pm0.002$ &$75.58\pm 1.40$ & --\\ 
                GINE \citep{hu2019strategies}                   &$0.252\pm 0.014$               & $0.088\pm0.002$ & $75.58\pm1.40$ &$86.68\pm 0.77$\\ 
                PNA \citep{corso2020principal}                  &$0.133\pm 0.011$               & $0.320\pm 0.032$ &$79.05\pm 1.32$& -- \\  
                HIMP \citep{fey2020hierarchical}                & $0.151\pm 0.002$              & $0.036\pm0.002$ &$78.80\pm0.82$ &$\mathbf{87.36\pm 0.50}$\\

                CIN \citep{bodnar2021weisfeiler}                 & $0.079\pm 0.006$              & $\mathbf{0.022\pm0.002}$ &$\mathbf{80.94\pm 0.57}$& -- \\

                DS-GNN (EGO+) \citep{bevilacqua2022equivariant} & $0.105\pm 0.003$              & -- &$77.40 \pm 2.19$ &$76.39\pm1.18$\\
                DSS-GNN (EGO+) \citep{bevilacqua2022equivariant}& $0.097\pm 0.006$              & -- &$76.78\pm 1.66$ &$77.95\pm0.40$\\
                GNN-AK+ \citep{zhao2021stars}                   & $0.091\pm 0.011$              & -- &$79.61\pm 1.19$ &--\\ 
                
                SUN (EGO+) \citep{frasca2022understanding}      & $0.084\pm 0.002$              & -- &$80.03\pm 0.55$ &--\\ 
                \hline
                \textbf{First order \Ptensors{} (our model)}                                   &$\mathbf{0.071\pm0.004}$       & $\mathbf{0.024\pm 0.001}$ & $\mathbf{80.76\pm 0.82}$ & $84.95\pm 0.58$\\\hline
        \end{tabular}
}
\vspace{-6pt}
        \caption{Experimental results on molecular datasets with baselines taken from \citep{frasca2022understanding,bodnar2021weisfeiler,bevilacqua2022equivariant}.
        }
\end{table*}

%% file: results.tex
\section{EXPERIMENTAL RESULTS}

We validated our model on several molecular datasets with one of the simplest possible realizations 
of the \Ptensors{} framework. The subgraphs are limited to vertices, edges and cycles (of any length) 
and the order of the corresponding neurons is either zero or one. 

The interactions between vertices and edges can be conceptualized as a classical MPNN: 
vertices send messages to edges, then messages are passed back to vertices, 
and vertices update from the incoming messages and their previous state. 
In this interaction, we are also able to get a partial update for the edges using the information from the vertices.
We similarly perform message passing between edges and cycles, with two notable distinctions: 
(1) since cycles are first order representations, we consider both internal linear maps when updating them, 
as apposed to the single linear map for vertex and edge updates; 
(2) We limit the edge-cycle interactions to edges that are fully encapsulated by 
cycles that they are interacting with. 
In some of our models we also used cycle-cycle interactions. 
In other respects, such as the placement of MLPs, etc., 
our model is similar to Cellular Isomorphism Networks (CIN) \citep{bodnar2021weisfeiler}. 
For details, please see the supplementary materials.

We report results on four molecular datasets that are standard benchmarks in the literature: 
(1) The full ZINC dataset of almost 250K organic molecules \citep{zinc15,gomez2018automatic}; 
(2) its more commonly used subset of just 
12K molecules; (3) the OGBG-MolHIV classification datatet of \m{~41K} molecules \cite{hu2020ogb}; 
(4) the TOX21 property prediction benchmark on 78311 molecules \citep{hu2019strategies}.
As baselines we use the best performing alternative algorithms from the literature of the same class, 
in particular, we do not compare to transformers or algorithms that explicitly take into account the 
3D positions of atoms. 

Remarkably, our \Ptensors{} based equivariant message passing algorithm is competitive with the 
best performing algorithms one each dataset, and beats all the other algorithms on ZINC 12K. 
We hypothesize that this is a direct result of the algorithm's greater expressivity. 
Additional experimental data can be found in the Supplementary Materials. 

%% file: conclusions.tex
\vspace{-6pt}
\section{CONCLUSIONS}
\vspace{-3pt}

The \m{P}-tensors framework unifies and generalizes equivariant message passing across a range of 
 subgraph neural network models, as well as some other models, such as certain simplicial complex 
neural networks, that are not based on subgraphs \emph{per se}, but ultimately still depend on the same 
equivariance constraints. The experimental results suggest that even in the first order case, 
the increased expressive power of our model helps improve upon the permformance of 
other graph neural networks on standard molecular benchmarks.

One of the advantages of our framework is that 
instead of implementing each type of  
subgraph interaction separately, in a piecemeal fashion, it makes it possible 
to formulate higher order message passing 
as a generic computational paradigm that is reusable across many models. 
In ongoing work we are developing a software library for higher order message passing 
that follows this philosophy.

%% file: checklist.tex
\section*{Checklist}

 \begin{enumerate}

 \item For all models and algorithms presented, check if you include:
 \begin{enumerate}
   \item A clear description of the mathematical setting, assumptions, algorithm, and/or model. [Yes]
   \item An analysis of the properties and complexity (time, space, sample size) of any algorithm. [Not Applicable]
   \item (Optional) Anonymized source code, with specification of all dependencies, including external libraries. [No]
 \end{enumerate}

 \item For any theoretical claim, check if you include:
 \begin{enumerate}
   \item Statements of the full set of assumptions of all theoretical results. [Yes]
   \item Complete proofs of all theoretical results. [No]
   \item Clear explanations of any assumptions. [Yes]     
 \end{enumerate}

 \item For all figures and tables that present empirical results, check if you include:
 \begin{enumerate}
   \item The code, data, and instructions needed to reproduce the main experimental results (either in the supplemental material or as a URL). [Yes]
   \item All the training details (e.g., data splits, hyperparameters, how they were chosen). [Yes]
         \item A clear definition of the specific measure or statistics and error bars (e.g., with respect to the random seed after running experiments multiple times). [Yes]
         \item A description of the computing infrastructure used. (e.g., type of GPUs, internal cluster, or cloud provider). [Yes]
 \end{enumerate}

 \item If you are using existing assets (e.g., code, data, models) or curating/releasing new assets, check if you include:
 \begin{enumerate}
   \item Citations of the creator If your work uses existing assets. [Yes]
   \item The license information of the assets, if applicable. [Not Applicable]
   \item New assets either in the supplemental material or as a URL, if applicable. [Yes]
   \item Information about consent from data providers/curators. [Not Applicable]
   \item Discussion of sensible content if applicable, e.g., personally identifiable information or offensive content. [Not Applicable]
 \end{enumerate}

 \item If you used crowdsourcing or conducted research with human subjects, check if you include:
 \begin{enumerate}
   \item The full text of instructions given to participants and screenshots. [Not Applicable]
   \item Descriptions of potential participant risks, with links to Institutional Review Board (IRB) approvals if applicable. [Not Applicable]
   \item The estimated hourly wage paid to participants and the total amount spent on participant compensation. [Not Applicable]
 \end{enumerate}

 \end{enumerate}

%% file: proofs2.tex

\setcounter{section}{0}
\section{Proofs}

\setcounter{theorem}{0}

\begin{theorem}
Any higher order MPNN in which 
\vspace{-6pt}
\begin{compactenum}[~(a)]
\item the subgraphs in each layer are selected using an invariant 
subgraph selection rule; 
\item the output of each subgraph neuron is a P-tensor; 
\item the messages sent from the \m{P}-tensors in each layer \m{\Fcal^{\textrm{in}}} 
to the \m{P}-tensors in following layer \m{\Fcal^{\textrm{out}}} are linear and satisfy the conditions of 
Definitions \ref{def: Ptensor-equi} and \ref{def: Ptensor-relabeling} 
\end{compactenum}
\vspace{-6pt}
is a permutation equivariant 
MPNN in the sense of Definition \ref{def: equivariant HOSNN}.
\end{theorem}

\begin{pfof}{Theorem 1}
Let \m{\Fcal^{\textrm{in}}=\cbrN{\Fin_1,\ldots,\Fin_m}} be a layer of \m{P}-tensors that sends messages to 
another layer \m{\Fcal^{\textrm{out}}=\cbrN{\Fout_1,\ldots,\Fout_{\wbar{m}}}}. 
Let \m{{\Fcal^{\textrm{in}}}{}'=\cbrN{\Fin_1{}',\ldots,\Fin_m{}'}} and
\m{{\Fcal^{\textrm{out}}}{}'=\cbrN{\Fout_1{}',\ldots,\Fout_{\wbar{m}}{}'}} be the corresponding layers 
after permuting the vertices of the underlying graph by some permutation \m{\sigma}. 

By the invariance of the subset selection rule, for each \m{\Fin_a} with reference domain 
\m{\Dcal=(x_1,\ldots,x_d)}, there is a corresponding \m{\Fin_{a'}{}'} with reference domain 
\m{\Dcal'=(\sigma(x_{\tau^{-1}(1)}),\ldots,\sigma(x_{\tau^{-1}(d)}))=\sigma\bullet\tau\circ\Dcal} for some permutation \m{\tau\tin\Sbb_{d}}. 
Similarly, for any \m{\Fout_{\wbar{a}}} with reference domain 
\sm{\wbar{\Dcal}=(\wbar{x}_1,\ldots,\wbar{x}_{\wbar{d}})}, 
there is a corresponding \sm{\Fout_{\wbar{a}'}{}'} with reference domain 
\sm{\wbar{\Dcal}'=(\sigma(\wbar{x}_{\wbar{\tau}^{-1}(1)}),\ldots,\sigma(x_{\wbar{\tau}^{-1}(\wbar{d})}))=\sigma\bullet\wbar{\tau}\circ\wbar{\Dcal}} 
for some \m{\wbar{\tau}\tin\Sbb_{\wbar{d}}}. 

Assuming that \m{\Fcal^{\textrm{in}}} is a covariant layer, its \m{P}-tensors before and after permutation 
are related by \m{\Fin_{a'}{}'=\tau\circ \Fin_a}. 
If the message passing process mapping \m{\Fcal^{\textrm{in}}\mapsto \Fcal^{\textrm{out}}}  
is relabeling invariant and permutation equivariant in the sense of 
Definitions \ref{def: Ptensor-equi} and \ref{def: Ptensor-relabeling},  
\[\Fout_{\wbar{a}'}{}'=\phi_{\Dcal',\wbar{\Dcal}'}(\Fin_{a'}{}')=
\phi_{\sigma\bullet\tau\circ\Dcal,\,\sigma\bullet\wbar{\tau}\circ\wbar{\Dcal}}(\Fin_{a'}{}')=
\wbar{\tau}\circ \phi_{\Dcal,\wbar{\Dcal}}(\Fin_{a})=\wbar{\tau}\circ\Fout_{\wbar{a}}\]
showing that \m{\Fcal^{\textrm{out}}} is also covariant. 
Since this relationship holds for any \m{(\Fcal^{\textrm{in}},\Fcal^{\textrm{out}})} pair of layers, 
the network as a whole is equivariant.  
\end{pfof}

%% file: appendix-experimental-details.tex
\section{Additional experiments}

Our experimental approach is designed to show how introducing higher order features can improve performence amongst a wide range of datasets.
To this end, we pick a model closely related to various ones found in the literature, 
but modified to included these extra features and make use of the given P-Tensor operations. 
In addition to the results presented in the main body of the paper we also ran 
experiments on the classic TU datasets (Table \ref{tbl: TU-datasets}).

\subsubsection*{Tested Variations}

\noindent
\textbf{Cycle-Cycle interactions.}
In addition to having first order and zeroth order interactions, we also considered cycle-cycle interactions, with their corrosponding five linear maps.
We mainly considered this on ZINC, MolTox21, and OGBG-MolHIV, but noticed a spike in validation volitility towards the end of training, damaging the reliablility for the model in many cases.
A similar issue came up in \citep{frasca2022understanding} for training on OGBG-MolHIV, and the issue was elieviated by using the ASAM optimizer \citep{kwon2021asam}, 
which is designed to reduce sharpness during training.
We found that this had limited success on the same dataset for our model.
Our hypothesis is that for highly irregular interactions, batch normalization struggles to converge because of the natural volitility that comes with such irregular structures.

\subsubsection*{Implementation}

\noindent
We implemented our algorithm in Pytorch \citep{NEURIPS2019_9015} using PyTorch Geometric \citep{Fey/Lenssen/2019}.
To add some encapsulation and improve runtime, we utilized pytorch lightning \citep{pytorchLightning} along with TensorBoard for visualization \citep{abadi2016tensorflow}.
For cycle finding, we implemented \citep{ferreira2014amortized}, and ran it along with all other structure map finding during preprocessing.
Thus, during runtime the only computations involving the structure of the graphs were the scatter operations themselves.
We ran our experiments on an NVIDIA GeForce GTX 1080. As for running time, on MolHIV a training runs averaged $69.3\pm 1.2$ minutes.
The source code of our implementation can be found at \href{https://github.com/arhands/ptensors}{\texttt{https://github.com/arhands/ptensors}}.

\subsection*{Hyperparameter Selection}

\noindent
Aside from some initial variations for designing our model, we limited our hyperparameter search space to primarily consider learning rates, the number of layers used, and the reductions used between cycles and edges.
Initial testing revealed high validation volitility, similar to that described in \citep{frasca2022understanding}, but we found that this was partially mitigated by reducing the momentum used in batch normalization, so we also considered that as part of our hyperparameters.

Our experimental setup for ZINC/ZINC-10K, and MolHIV is based on \citep{bodnar2021weisfeiler,Sardellitti2021}, while our setup for Tox21 is based on \citep{fey2020hierarchical}.
We base our experimental setup for the TUDatasets on \citep{frasca2022understanding}, utilizing the same hyperparameter grid and base node/edge embeddings.

\input{tbl-tu-datasets}

\section{Elementary discussion of our models}

For completeness in the following we give an elementary description of one of our models, showing 
how most P-tensor operations, at least in the first order case, can be implemented with customary GNN operations. 

We will use $\mu_i$ to denote a generic multilayer perceptrion (MLP) with the input parameters concatenated together channel-wise.
In our experiments, each linear layer is followed by a batch normalization layer \citep{ioffe2015batch} and then a ReLU activation, unless specified otherwise.
We also set multilayer MLPs to have double the hidden channels within their respective internal layers.
Let $\Gcal = (V,E)$ be a simple undirected graph.
Let $X^v$ and $X^e$ be a zeroth order representation of the vertices and edges in $\Gcal$, respectively. 
To help simplify notation, we shall consider a given edge $e_{i,j} = \{i,j\} = e_{j,i}$.

Our model can be then be described as a sequence of layers, each allowing for interactions between subgraphs of $\Gcal$. To begin with, we shall define how cycles and vertices interact.
Then, in each layer of our model, we capture the expressive power of MPNNs by transferring from vertices to edges, and then back to vertices.
\begin{align}
    Y^v_{i}
    &= \mu_1\left(X^v_i,\sum_{e_{i j}\in E}\mu_2\left(X^e_{e_{i,j}},X^v_i + X^v_j\right) \right)\label{eqn:exp setup 1.1.1}
    \intertext{Where $i\in V$, $\mu_1$ contains two layers, and $\mu_2$ has a single layer. 
    As can be seen in the above equation, the inner sum corresponds to the single linear map from vertices to edges, and the outer sum corresponds to the other direction.
    Similarly, we get the incoming messages from vertices to edges using a single transfer operation.}
    Y^{v\rightarrow e}_{e_{i,j}}
    &= \mu_3\left(X^e_{e_{i,j}},X^v_i + X^v_j\right)\label{eqn:exp setup 1.1.2}
    \intertext{So far, this can largely be seen as classical message passing. However, where things get more interesting when we consider the interactions between edges. 
    Let $X^c$ denote a representation of selected cycles on $\Gcal$ (see hyperparameter subsection for per-dataset details) and, in similar fashion to \citep{bodnar2021weisfeiler}, 
    let $\beta_{\downarrow}(e)$ denote the cycles that contain a given edge $e$ and $\beta_{\uparrow}(c)$ be the edges covered entirely by a given cycle $c$.
    Edges interact with cycles in a way analogous to vertices to edges, with the key difference being cycles use a higher order representation. 
    First, consider the following equation depicting how edges send messages to cycles, denoting $H^{e\rightarrow c}_{c,i}$ as the value received at index $i\in V\cap c$ 
    for the first order P-tensor corresponding to a given cycle $c$.}
    H^{e\rightarrow c}_{c,i}
    &= 
    \sum_{\substack{e\in\beta_{\uparrow}(c)\\ i:\in e}} X^e_e
    \conc
    \sum_{e\in\beta_{\uparrow}(c)} X^e_e
    \label{eqn:exp setup 1.1.3}
    \intertext{
        This can be equated to the first summation in equation \ref{eqn:exp setup 1.1.1}, but where we gain an additional linear map by considering the overlapping vertices separately.
        It is worth noting for completeness that in practice we considered both mean and sum reductions for the summations in \ref{eqn:exp setup 1.1.3}.
        We can then get the update to a given cycle by combining the incoming representation with the representation obtained by computing the internal linear maps from $X^c_c$ to itself.
    }
    Y^{c}_{c,i}
    &= \mu_4\left((1 + \varepsilon_{1})\left(X^c_{c,i}\conc\sum_{j\in c} X^c_{c,j}\right) + H^{e\rightarrow c}_{c,i}\right)
    \label{eqn:exp setup 1.1.4}
    \intertext{Where $\mu_4$ has two layers and $\varepsilon_1$ is a learnable scalar. 
    We also use $H^{e\rightarrow c}$ to compute messages sent back to edges in a similar way to equation \ref{eqn:exp setup 1.1.1}.
    }
    Y^{c\rightarrow e}_{e}
    &= \mu_5\left(
        (1 + \varepsilon_{2}) Y^{e\rightarrow v}_{e} +
        (1 + \varepsilon_{3}) \sum_{c\in\beta_{\downarrow}(e)}\sum_{i\in e}  \mu_6\left(H^c_{c,i},X^c_{c,i}\right) +
        \sum_{c\in\beta_{\downarrow}(e)}\sum_{i\in c}  \mu_6\left(H^c_{c,i},X^c_{c,i}\right)
    \right)\label{eqn:exp setup 1.1.5}
    \intertext{
        Where $\mu_5$ has two layers, $\mu_6$ has one, and $\varepsilon_2$ and $\varepsilon_3$ are learnable weights.
        Here we can see a direct analogy between this and equation \ref{eqn:exp setup 1.1.1}, where $\mu_5$ corresponds to $\mu_1$ and $\mu_6$ corresponds to $\mu_2$.
        It is again worth noting that there are many options for the given reductions, but we simply least summations for ease of notation.
        Since edges receive messages from both vertices and cycles, we compute their new state via a single layer perceptron, $\mu_7$:
    }
    Y^e_e
    &= \mu_7\left(Y^{v\rightarrow e}_{e},Y^{c\rightarrow e}_{e}\right)
\end{align}

%% file: tbl-tu-datasets.tex
\newcommand{\first}[1]{\color{red}{#1}}
\newcommand{\second}[1]{\color{purple}{#1}}
\newcommand{\third}[1]{\ifmmode\mathbf{#1}\else\textbf{#1}\fi}
\begin{table}[t]
    \centering
    {\tiny
    \begin{tabular}{l|llll|ll}
    \hline
    Dataset                                                             & MUTAG                     & PTC          & PROTEINS      & NCI1         & IMDB-B       & IMDB-M       \\ \hline
    DCNN\citep{atwood2016diffusion}                                     & --                        & --           & $61.3\pm1.6$  & $56.6\pm1.0$ & $49.1\pm1.4$ & $33.5\pm1.4$ \\
    DGCNN\citep{zhang2018end}                                           & $85.8\pm1.8$              & $58.6\pm2.5$ & $75.5\pm0.9$  & $74.4\pm0.5$ & $70.0\pm0.9$ & $47.8\pm0.9$ \\
    IGN\citep{KerivenPeyre2019}                                         & $83.9\pm13.0$             & $58.5\pm6.9$ & $76.6\pm5.5$  & $74.3\pm2.7$ & $72.0\pm5.5$ & $48.7\pm3.4$ \\
    PPGNs\citep{maron2019provably}                                      & $90.6\pm8.7$              & $66.2\pm6.6$ & $\first{77.2\pm4.7}$  & $83.2\pm1.1$ & $73.0\pm5.8$ & $50.5\pm3.6$ \\
    Natural GN\citep{de2020natural}                                     & $89.4\pm1.6$              & $66.8\pm1.7$ & $71.7\pm1.0$  & $82.4\pm1.3$ & $73.5\pm2.0$ & $51.3\pm1.5$ \\
    GSN\citep{hu2019gsn}                                    & $\third{92.2\pm7.5}$      & $\third{68.2\pm7.2}$ & $76.6\pm5.0$  & $83.5\pm2.0$ & $\second{77.8\pm3.3}$ & $\first{54.3\pm3.3}$ \\
    SIN\citep{pmlr-v139-bodnar21a}                                     & --                        & --           & $76.4\pm3.3$  & $82.7\pm2.1$ & $75.6\pm3.2$ & $52.7\pm3.1$ \\
    CIN\citep{bodnar2021weisfeiler}                                      & $\second{92.7\pm6.1}$     & $\third{68.2\pm5.6}$ & $\third{77.0\pm4.3}$  & $83.6\pm1.4$ & $75.6\pm3.7$ & $52.7\pm3.1$ \\ \hline
    GIN\citep{xu2018powerful}                                           & $89.4\pm5.6$              & $64.6\pm7.0$ & $76.2\pm2.8$  & $82.7\pm1.7$ & $75.1\pm5.1$ & $52.3\pm2.8$ \\ \hline
    GIN + ID-GNN\citep{you2021identity}                                 & $90.4\pm5.6$              & $67.2\pm4.3$ & $75.4\pm2.7$  & $82.6\pm1.6$ & $76.0\pm2.7$ & $52.7\pm4.2$ \\
    DropEdge\citep{rong2019dropedge}                                    & $91.0\pm5.7$              & $64.5\pm2.6$ & $73.5\pm4.5$  & $82.0\pm2.6$ & $76.5\pm3.3$ & $52.8\pm2.8$ \\ \hline
    DS-GNN (GIN) (ND)\citep{bevilacqua2022equivariant}                  & $89.4\pm4.8$              & $66.3\pm7.0$ & $\second{77.1\pm4.6}$  & $83.8\pm2.4$ & $75.4\pm2.9$ & $52.7\pm2.0$ \\
    DS-GNN (GIN) (EGO)\citep{bevilacqua2022equivariant}                 & $89.9\pm6.5$              & $68.6\pm5.8$ & $76.7\pm5.8$  & $81.4\pm0.7$ & $76.1\pm2.8$ & $52.6\pm2.8$ \\
    DS-GNN (GIN) (EGO+) \citep{bevilacqua2022equivariant}               & $91.1\pm7.0$              & $69.2\pm6.5$ & $75.9\pm4.3$  & $83.7\pm1.8$ & $\third{77.1\pm3.0}$ & $53.2\pm2.8$ \\ \hline
    DSS-GNN (GIN) (ND)  \citep{bevilacqua2022equivariant}               & $91.0\pm3.5$              & $66.3\pm5.9$ & $76.1\pm3.4$  & $83.6\pm1.5$ & $76.1\pm2.9$ & $\second{53.3\pm1.9}$ \\
    DSS-GNN (GIN) (EGO) \citep{bevilacqua2022equivariant}               & $91.0\pm4.7$              & $68.2\pm5.8$ & $76.7\pm4.1$  & $83.6\pm1.8$ & $76.5\pm2.8$ & $\second{53.3\pm3.1}$ \\
    DSS-GNN (GIN) (EGO+) \citep{bevilacqua2022equivariant}              & $91.1\pm7.0$              & $\second{69.2\pm6.5}$ & $75.9\pm4.3$  & $83.7\pm1.8$ & $\third{77.1\pm3.0}$ & $\third{53.2\pm2.4}$ \\ \hline
    GIN-AK+              \citep{zhao2021stars}                          & $91.3\pm7.0$              & $67.8\pm8.8$ & $\second{77.1\pm5.7}$  & $\first{85.0\pm2.0}$ & $75.0\pm4.2$ & --           \\ \hline
    SUN (GIN) (NULL)\citep{frasca2022understanding}                     & $91.6\pm4.8$              & $67.5\pm6.8$ & $76.8\pm4.4$  & $84.1\pm2.0$ & $76.2\pm1.9$ & $52.6\pm3.2$ \\
    SUN (GIN) (NM)   \citep{frasca2022understanding}                    & $91.0\pm4.7$              & $67.0\pm4.8$ & $75.7\pm3.4$  & $\second{84.2\pm1.5}$ & $76.1\pm2.9$ & $53.1\pm2.5$ \\
    SUN (GIN) (EGO)  \citep{frasca2022understanding}                    & $\second{92.7\pm5.8}$     & $67.2\pm5.9$ & $76.8\pm5.0$  & $83.7\pm1.3$ & $76.6\pm3.4$ & $52.7\pm2.3$ \\
    SUN (GIN) (EGO+) \citep{frasca2022understanding}                    & $92.1\pm5.8$              & $67.6\pm5.5$ & $76.1\pm5.1$  & $\second{84.2\pm1.5}$ & $76.3\pm1.9$ & $52.9\pm2.8$ \\ \hline
    
    Ours                                                                & $\first{92.9\pm1.7}$      & $\first{71.7\pm5.2}$ & $75.9\pm2.5$  & $\second{84.2\pm1.7}$ & $\first{77.9\pm3.2}$ & $\first{54.3\pm2.0}$ \\ \hline
    \end{tabular}
    }
    \caption{\label{tbl: TU-datasets}
        Summary of results on TUDatasets with baselines taken from \citep{frasca2022understanding}. Top three scores are given in \first{red}, \second{purple}, and \third{bold}.
    }
\end{table}

%% file: bibliography.bib
@article{corso2020principal,
	title        = {Principal neighbourhood aggregation for graph nets},
	author       = {Corso, Gabriele and Cavalleri, Luca and Beaini, Dominique and Li{\`o}, Pietro and Veli{\v{c}}kovi{\'c}, Petar},
	year         = 2020,
	journal      = {Advances in Neural Information Processing Systems},
	volume       = 33,
	
}

@inproceedings{kipf17,
	title        = {Semi-Supervised Classification with Graph Convolutional Networks},
	author       = {Thomas N. Kipf and Max Welling},
	year         = 2017,
	booktitle    = {International Conference on Learning Representations},
	url          = {https://openreview.net/forum?id=SJU4ayYgl},
}

@inproceedings{xu2018powerful,
title={How Powerful are Graph Neural Networks?},
author={Keyulu Xu and Weihua Hu and Jure Leskovec and Stefanie Jegelka},
booktitle={International Conference on Learning Representations},
year={2019},
}

@inproceedings{you2021identity,
  title={Identity-aware graph neural networks},
  author={You, Jiaxuan and Gomes-Selman, Jonathan M and Ying, Rex and Leskovec, Jure},
  booktitle={Proceedings of the AAAI conference on artificial intelligence},
  volume={35},
  number={12},
  year={2021}
}

@inproceedings{rong2019dropedge,
title={DropEdge: Towards Deep Graph Convolutional Networks on Node Classification},
author={Yu Rong and Wenbing Huang and Tingyang Xu and Junzhou Huang},
booktitle={International Conference on Learning Representations},
year={2020},
url={https://openreview.net/forum?id=Hkx1qkrKPr}
}

@inproceedings{maron2018,
title={Invariant and Equivariant Graph Networks},
author={Haggai Maron and Heli Ben-Hamu and Nadav Shamir and Yaron Lipman},
booktitle={International Conference on Learning Representations},
year={2019},
url={https://openreview.net/forum?id=Syx72jC9tm},
}

@inproceedings{Sardellitti2021,
	title        = {Topological Signal Processing over Cell Complexes},
	author       = {Sardellitti, Stefania and Barbarossa, Sergio and Testa, Lucia},
	year         = 2021,
	booktitle    = {2021 55th Asilomar Conference on Signals, Systems, and Computers},
	
	doi          = {10.1109/IEEECONF53345.2021.9723256},
}

@inproceedings{pmlr-v139-bodnar21a,
  title={{W}eisfeiler and {L}ehman go Topological: Message Passing Simplicial Networks},
  author={Bodnar, Cristian and Frasca, Fabrizio and Wang, Yuguang and Otter, Nina and Montufar, Guido F and Lio, Pietro and Bronstein, Michael},
  booktitle={International Conference on Machine Learning},
  year={2021},
}

@inproceedings{murphy2019relational,
  title={Relational pooling for graph representations},
  author={Murphy, Ryan and Srinivasan, Balasubramaniam and Rao, Vinayak and Ribeiro, Bruno},
  booktitle={International Conference on Machine Learning},
  year={2019},
}

@inproceedings{hu2019strategies,
title={Strategies for Pre-training Graph Neural Networks},
author={Hu, Weihua and Liu, Bowen and Gomes, Joseph and Zitnik, Marinka and Liang, Percy and Pande, Vijay and Leskovec, Jure},
booktitle={International Conference on Learning Representations},
year={2020},
url={https://openreview.net/forum?id=HJlWWJSFDH},
}

@inproceedings{ioffe2015batch,
  title={Batch normalization: Accelerating deep network training by reducing internal covariate shift},
  author={Ioffe, Sergey and Szegedy, Christian},
  booktitle={International conference on machine learning},
  year={2015},
}

@article{bodnar2021weisfeiler,
  title={{W}eisfeiler and {L}ehman go Cellular: {CW} Networks},
  author={Bodnar, Cristian and Frasca, Fabrizio and Otter, Nina and Wang, Yuguang and Lio, Pietro and Montufar, Guido F and Bronstein, Michael},
  journal={Advances in Neural Information Processing Systems},
  volume={34},
  pages={2625--2640},
  year={2021}
}

@inproceedings{pmlr-v97-maron19a,
	title        = {On the universality of invariant networks},
	author       = {Maron, Haggai and Fetaya, Ethan and Segol, Nimrod and Lipman, Yaron},
	year         = 2019,
	booktitle    = {International conference on machine learning},
}

@article{maron2019provably,
	title        = {Provably powerful graph networks},
	author       = {Maron, Haggai and Ben-Hamu, Heli and Serviansky, Hadar and Lipman, Yaron},
	year         = 2019,
	journal      = {Advances in Neural Information Processing Systems},
	volume       = 32,
}

@inproceedings{zhang2018end,
  title={An end-to-end deep learning architecture for graph classification},
  author={Zhang, Muhan and Cui, Zhicheng and Neumann, Marion and Chen, Yixin},
  booktitle={Proceedings of the AAAI conference on artificial intelligence},
  year={2018}
}

@article{de2020natural,
  title={Natural graph networks},
  author={de Haan, Pim and Cohen, Taco S and Welling, Max},
  journal={Advances in Neural Information Processing Systems},
  volume={33},
  pages={3636--3646},
  year={2020}
}

@article{atwood2016diffusion,
  title={Diffusion-convolutional neural networks},
  author={Atwood, James and Towsley, Don},
  journal={Advances in Neural Information Processing Systems},
  volume={29},
  year={2016}
}

@inproceedings{bevilacqua2022equivariant,
	title        = {Equivariant Subgraph Aggregation Networks},
	author       = {Beatrice Bevilacqua and Fabrizio Frasca and Derek Lim and Balasubramaniam Srinivasan and Chen Cai and Gopinath Balamurugan and Michael M. Bronstein and Haggai Maron},
	year         = 2022,
	booktitle    = {International Conference on Learning Representations},
}

@inproceedings{zhao2021stars,
	title        = {From Stars to Subgraphs: Uplifting Any {GNN} with Local Structure Awareness},
	author       = {Lingxiao Zhao and Wei Jin and Leman Akoglu and Neil Shah},
	year         = 2022,
	booktitle    = {International Conference on Learning Representations},
	url          = {https://openreview.net/forum?id=Mspk_WYKoEH},
}

@inproceedings{gilmer2017neural,
	title        = {Neural message passing for quantum chemistry},
	author       = {Gilmer, Justin and Schoenholz, Samuel S and Riley, Patrick F and Vinyals, Oriol and Dahl, George E},
	year         = 2017,
	booktitle    = {International conference on machine learning},
}

@inproceedings{morris2019weisfeiler,
	title        = {{W}eisfeiler and {L}eman go neural: Higher-order graph neural networks},
	author       = {Morris, Christopher and Ritzert, Martin and Fey, Matthias and Hamilton, William L and Lenssen, Jan Eric and Rattan, Gaurav and Grohe, Martin},
	year         = 2019,
	booktitle    = {Proceedings of the AAAI conference on artificial intelligence},
}

@article{chen2020can,
	title        = {Can graph neural networks count substructures?},
	author       = {Chen, Zhengdao and Chen, Lei and Villar, Soledad and Bruna, Joan},
	year         = 2020,
	journal      = {Advances in Neural Information Processing Systems},
	volume       = 33,
}

@article{weisfeiler1968reduction,
	title        = {The reduction of a graph to canonical form and the algebra which appears therein},
	author       = {Weisfeiler, Boris and Leman, Andrei},
	year         = 1968,
	journal      = {NTI, Series},
	pages        = {12--16},
}

@inproceedings{zaheer2017deep,
author = {Zaheer, Manzil and Kottur, Satwik and Ravanbhakhsh, Siamak and P\'{o}czos, Barnab\'{a}s and Salakhutdinov, Ruslan and Smola, Alexander J},
title = {Deep Sets},
year = {2017},
booktitle = {Proceedings of the 31st International Conference on Neural Information Processing Systems},
}

@inproceedings{fey2020hierarchical,
  title={Hierarchical Inter-Message Passing for Learning on Molecular Graphs},
  author={Fey, M. and Yuen, J. G. and Weichert, F.},
  booktitle={ICML Graph Representation Learning and Beyond (GRL+) Workhop},
  year={2020},
}

@article{hu2020ogb,
	title        = {Open graph benchmark: Datasets for machine learning on graphs},
	author       = {Hu, Weihua and Fey, Matthias and Zitnik, Marinka and Dong, Yuxiao and Ren, Hongyu and Liu, Bowen and Catasta, Michele and Leskovec, Jure},
	year         = 2020,
	journal      = {Advances in Neural Information Processing Systems},
	volume       = 33,
}

@article{thiede2021,
	title        = {Autobahn: Automorphism-based graph neural nets},
	author       = {Thiede, Erik and Zhou, Wenda and Kondor, Risi},
	year         = 2021,
	journal      = {Advances in Neural Information Processing Systems},
	volume       = 34,
}

@article{frasca2022understanding,
	title        = {Understanding and Extending Subgraph {GNN}s by Rethinking Their Symmetries},
	author       = {Frasca, Fabrizio and Bevilacqua, Beatrice and Bronstein, Michael and Maron, Haggai},
	year         = 2022,
	journal      = {Advances in Neural Information Processing Systems},
	volume       = 35,
}

@inproceedings{kwon2021asam,
  title={{Asam}: Adaptive sharpness-aware minimization for scale-invariant learning of deep neural networks},
  author={Kwon, Jungmin and Kim, Jeongseop and Park, Hyunseo and Choi, In Kwon},
  booktitle={International Conference on Machine Learning},
  year={2021},
}

@incollection{NEURIPS2019_9015,
	title        = {Py{T}orch: An Imperative Style, High-Performance Deep Learning Library},
	author       = {Paszke, Adam and Gross, Sam and Massa, Francisco and Lerer, Adam and Bradbury, James and Chanan, Gregory and Killeen, Trevor and Lin, Zeming and Gimelshein, Natalia and Antiga, Luca and Desmaison, Alban and Kopf, Andreas and Yang, Edward and DeVito, Zachary and Raison, Martin and Tejani, Alykhan and Chilamkurthy, Sasank and Steiner, Benoit and Fang, Lu and Bai, Junjie and Chintala, Soumith},
	year         = 2019,
	booktitle    = {Advances in Neural Information Processing Systems 32},
	url          = {http://papers.neurips.cc/paper/9015-pytorch-an-imperative-style-high-performance-deep-learning-library.pdf},
}

@inproceedings{Fey/Lenssen/2019,
	title        = {Fast Graph Representation Learning with {PyTorch Geometric}},
	author       = {Fey, Matthias and Lenssen, Jan E.},
	year         = 2019,
	booktitle    = {ICLR Workshop on Representation Learning on Graphs and Manifolds},
}

@book{Sagan,
  title = {The Symmetric Group},
  ISBN = {9781475768046},
  ISSN = {0072-5285},
  url = {http://dx.doi.org/10.1007/978-1-4757-6804-6},
  DOI = {10.1007/978-1-4757-6804-6},
  journal = {Graduate Texts in Mathematics},
  publisher = {Springer New York},
  author = {Sagan,  Bruce E.},
  year = {2001}
}

@article{Sannai2019universal,
  author    = {Akiyoshi Sannai and
               Yuuki Takai and
               Matthieu Cordonnier},
  title     = {Universal approximations of permutation invariant/equivariant functions
               by deep neural networks},
  journal   = {CoRR},
  volume    = {abs/1903.01939},
  year      = {2019},
  archivePrefix = {arXiv},
  eprint    = {1903.01939}
}

@incollection{KerivenPeyre2019,
title = {Universal Invariant and Equivariant Graph Neural Networks},
author = {Keriven, Nicolas and Peyr\'{e}, Gabriel},
booktitle = {Advances in Neural Information Processing Systems 32},
year = {2019},
}

@misc{abadi2016tensorflow,
title={ {TensorFlow}: Large-Scale Machine Learning on Heterogeneous Systems},
url={https://www.tensorflow.org/},
note={Software available from tensorflow.org},
author={
    Mart\'{i}n~Abadi and
    Ashish~Agarwal and
    Paul~Barham and
    Eugene~Brevdo and
    Zhifeng~Chen and
    Craig~Citro and
    Greg~S.~Corrado and
    Andy~Davis and
    Jeffrey~Dean and
    Matthieu~Devin and
    Sanjay~Ghemawat and
    Ian~Goodfellow and
    Andrew~Harp and
    Geoffrey~Irving and
    Michael~Isard and
    Yangqing Jia and
    Rafal~Jozefowicz and
    Lukasz~Kaiser and
    Manjunath~Kudlur and
    Josh~Levenberg and
    Dandelion~Man\'{e} and
    Rajat~Monga and
    Sherry~Moore and
    Derek~Murray and
    Chris~Olah and
    Mike~Schuster and
    Jonathon~Shlens and
    Benoit~Steiner and
    Ilya~Sutskever and
    Kunal~Talwar and
    Paul~Tucker and
    Vincent~Vanhoucke and
    Vijay~Vasudevan and
    Fernanda~Vi\'{e}gas and
    Oriol~Vinyals and
    Pete~Warden and
    Martin~Wattenberg and
    Martin~Wicke and
    Yuan~Yu and
    Xiaoqiang~Zheng},
  year={2015},
}

@inproceedings{ferreira2014amortized,
  title={Amortized-delay algorithm for listing chordless cycles in undirected graphs},
  author={Ferreira, Rui and Grossi, Roberto and Rizzi, Romeo and Sacomoto, Gustavo and Sagot, Marie-France},
  booktitle={European Symposium on Algorithms},
  pages={418--429},
  year={2014},
  organization={Springer}
}

@misc{pytorchLightning,
author = {Falcon, William and {The PyTorch Lightning team}},
doi = {10.5281/zenodo.3828935},
license = {Apache-2.0},
month = mar,
title = {{PyTorch Lightning}},
url = {https://github.com/Lightning-AI/lightning},
version = {1.4},
year = {2019}
}

@article{gomez2018automatic,
  title={Automatic chemical design using a data-driven continuous representation of molecules},
  author={G{\'o}mez-Bombarelli, Rafael and Wei, Jennifer N and Duvenaud, David and Hern{\'a}ndez-Lobato, Jos{\'e} Miguel and S{\'a}nchez-Lengeling, Benjam{\'\i}n and Sheberla, Dennis and Aguilera-Iparraguirre, Jorge and Hirzel, Timothy D and Adams, Ryan P and Aspuru-Guzik, Al{\'a}n},
  journal={ACS central science},
  volume={4},
  number={2},
  pages={268--276},
  year={2018},
  publisher={ACS Publications}
}

@article{zinc15,
author = {Sterling, Teague and Irwin, John J.},
title = {{ZINC} 15 – {L}igand Discovery for Everyone},
journal = {Journal of Chemical Information and Modeling},
volume = {55},
number = {11},
pages = {2324-2337},
year = {2015},
doi = {10.1021/acs.jcim.5b00559},
    note ={PMID: 26479676},

URL = { 
    
        https://doi.org/10.1021/acs.jcim.5b00559
    
},
eprint = { 
    
        https://doi.org/10.1021/acs.jcim.5b00559
    
}

}


%% file: gnn1.bib
@string{icml = {Proceedings of International Conference on Machine Learning (ICML)}}

@string{nips = {Advances in Neural Information Processing Systems (NIPS)}}

@string{iclr = {International Conference on Learning Representations (ICLR)}}

@article{CCN-JCP,
author = {Hy,Truong Son  and Trivedi,Shubhendu  and Pan,Horace  and Anderson,Brandon M.  and Kondor,Risi },
title = {Predicting molecular properties with covariant compositional networks},
journal = {The Journal of Chemical Physics},
volume = {148},
number = {24},
pages = {241745},
year = {2018},
doi = {10.1063/1.5024797}
}

@article{thiede2020general,
  author       = {Erik Henning Thiede and
                  Truong{-}Son Hy and
                  Risi Kondor},
  title        = {The general theory of permutation equivarant neural networks and higher
                  order graph variational encoders},
  journal      = {CoRR},
  volume       = {abs/2004.03990},
  year         = {2020},
  url          = {https://arxiv.org/abs/2004.03990},
  eprinttype    = {arXiv},
  eprint       = {2004.03990},
  bibsource    = {dblp computer science bibliography, https://dblp.org}
}

@article{alsentzer2020subgraph,
  title={Subgraph neural networks},
  author={Alsentzer, Emily and Finlayson, Samuel and Li, Michelle and Zitnik, Marinka},
  journal={Advances in Neural Information Processing Systems},
  volume={33},
  pages={8017--8029},
  year={2020},
  series={NIPS'20}
}

@article{Feng2019,
archivePrefix = {arXiv},
arxivId = {1809.09401},
author = {Feng, Yifan and You, Haoxuan and Zhang, Zizhao and Ji, Rongrong and Gao, Yue},
doi = {10.1609/aaai.v33i01.33013558},
eprint = {1809.09401},
isbn = {9781577358091},
issn = {2159-5399},
journal = {33rd AAAI Conference on Artificial Intelligence},
title = {{Hypergraph neural networks}},
year = {2019}
}

@inproceedings{Ebli2020,
  title = {Simplicial Neural Networks},
  author = {Ebli, Stefania and Defferrard, Michaël and Spreemann, Gard},
  booktitle = {Topological Data Analysis and Beyond workshop at NeurIPS},
  year = {2020},
  archiveprefix = {arXiv},
  eprint = {2010.03633},
}

@article{dong2020hnhn,
  author       = {Yihe Dong and
                  Will Sawin and
                  Yoshua Bengio},
  title        = {{HNHN:} Hypergraph Networks with Hyperedge Neurons},
  journal      = {CoRR},
  volume       = {abs/2006.12278},
  year         = {2020},
  url          = {https://arxiv.org/abs/2006.12278},
  eprinttype    = {arXiv},
  eprint       = {2006.12278},
  bibsource    = {dblp computer science bibliography, https://dblp.org}
}

@article{hu2019gsn,
  author       = {Wenpeng Hu and
                  Zhangming Chan and
                  Bing Liu and
                  Dongyan Zhao and
                  Jinwen Ma and
                  Rui Yan},
  title        = {{GSN:} {A} Graph-Structured Network for Multi-Party Dialogues},
  journal      = {CoRR},
  volume       = {abs/1905.13637},
  year         = {2019},
  url          = {http://arxiv.org/abs/1905.13637},
  eprinttype    = {arXiv},
  eprint       = {1905.13637},
  bibsource    = {dblp computer science bibliography, https://dblp.org}
}
